\documentclass[runningheads]{llncs}
 
\usepackage[mobile]{eccv}

\usepackage{eccvabbrv}

\usepackage{graphicx}
\usepackage{booktabs}
\usepackage{epstopdf}
\usepackage{multirow}
\usepackage{xstring}
\usepackage{arydshln}
\usepackage{float}
\usepackage{amsmath}
\usepackage{flushend}
\usepackage{caption}
\usepackage{cuted}
\usepackage{colortbl}
\usepackage[normalem]{ulem}
\usepackage{marvosym}
\useunder{\uline}{\ul}{}

\usepackage[accsupp]{axessibility}  

\makeatletter
\newcommand{\printfnsymbol}[1]{%
  \textsuperscript{\@fnsymbol{#1}}%
}
\makeatother

\usepackage{hyperref}

\usepackage{orcidlink}

\title{SDPT: Synchronous Dual Prompt Tuning for Fusion-based Visual-Language Pre-trained Models} 

\titlerunning{SDPT: Synchronous Dual Prompt Tuning for Fusion-based VLPMs}

\author{Yang Zhou\inst{1}\orcidlink{0000-0003-2848-7642}\thanks{Equal contribution. \printfnsymbol{4} Corresponding author.} \and
Yongjian Wu\inst{1}\orcidlink{0009-0000-5631-164X}\printfnsymbol{1}\and
Jiya Saiyin\inst{1}\orcidlink{0009-0006-7212-6595} \and
Bingzheng Wei\inst{2}\orcidlink{0000-0001-6979-0459} \and
Maode Lai\inst{3} \and
Eric Chang\inst{4} \and
Yan Xu\inst{1}\orcidlink{0000-0002-2636-7594}\printfnsymbol{4}}

\authorrunning{Y. Zhou et al.}

\institute{School of Biological Science and Medical Engineering, Beihang University \and ByteDance Inc. \and Zhejiang University \and Taiwan Artificial Intelligence Foundation\\
\email{\{zhouyangbme,wuyongjian\}@buaa.edu.cn}\;\;\;  \email{xuyan04@gmail.com}}

\begin{document}
\maketitle

\begin{abstract}
Prompt tuning methods have achieved remarkable success in parameter-efficient fine-tuning on large pre-trained models. However, their application to dual-modal fusion-based visual-language pre-trained models (VLPMs), such as GLIP, has encountered issues. Existing prompt tuning methods have not effectively addressed the modal mapping and aligning problem for tokens in different modalities, leading to poor transfer generalization. To address this issue, we propose Synchronous Dual Prompt Tuning (SDPT). SDPT initializes a single set of learnable unified prototype tokens in the established modal aligning space to represent the aligned semantics of text and image modalities for downstream tasks. Furthermore, SDPT establishes inverse linear projections that require no training to embed the information of unified prototype tokens into the input space of different modalities. The inverse linear projections allow the unified prototype token to synchronously represent the two modalities and enable SDPT to share the unified semantics of text and image for downstream tasks across different modal prompts. Experimental results demonstrate that SDPT assists fusion-based VLPMs to achieve superior outcomes with only 0.04\% of model parameters for training across various scenarios, outperforming other single- or dual-modal methods. The code will be released at \href{https://github.com/wuyongjianCODE/SDPT}{https://github.com/wuyongjianCODE/SDPT}.
\end{abstract}

\section{Introduction}
Recently, visual-language pre-trained models (VLPMs) have significantly advanced computer vision tasks \cite{chen2023vlp, patashnik2021styleclip,li2022blip,jain2021putting}, and one particular type of VLPM is fusion-based \cite{li2022grounded, zhang2022glipv2, dou2022coarse, li2023gligen}. Fusion-based VLPMs, typified by GLIP \cite{li2022grounded}, elevate visual-language representation learning to object-level by pre-training with grounding data and introducing multi-layer cross multi-head attentions (X-MHA) between image and text encoders as bimodal deep fusion. Despite that fusion-based VLPMs achieve considerable zero-shot detection or grounding performance through prompts, fine-tuning is still necessary for specific downstream tasks, especially when target category texts and images are scarce in the pre-training data. However, full fine-tuning on downstream tasks is hindered by the massive size of these models \cite{zhuang2020comprehensive}. Parameter-efficient fine-tuning (PEFT) is a potential solution to address this challenge \cite{he2021towards}.

\begin{figure}[t]
\centering
\includegraphics[width=\columnwidth]{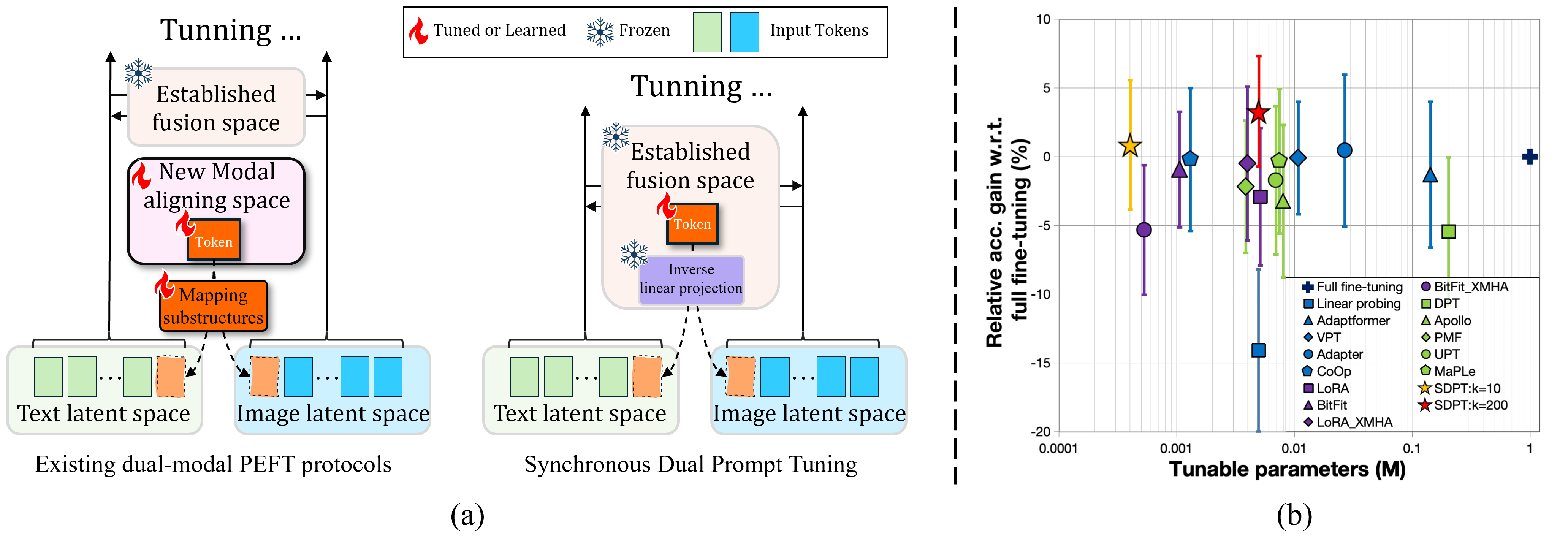}
  \caption{Synchronous Dual Prompt Tuning (SDPT) \textit{vs.} other dual-modal PEFT methods on fusion-based VLPMs. (a) Existing dual-modal PEFT methods (left) require learning new modal mapping substructures or modality aligning spaces, whereas SDPT (right) does not and thus achieves better PEFT performance for fusion-based VLPMs on new tasks. (b) Performance of different methods on 13 downstream tasks in ODinW13 \cite{li2022elevater} for GLIP-L, with mean and standard deviation annotated. SDPT (k=10) outperforms full fine-tuning while using only 0.04\% of all model parameters.}
  \label{briefcompare}
\end{figure}

Prompt tuning (PT) methods, compatible with transformer architectures and highly effective, have emerged as prevalent PEFT approaches. These methods add learnable tokens to input, expecting these learnable tokens to incorporate knowledge from the downstream task \cite{lester2021power,zhou2022learning}. However, many PT methods are primarily developed for unimodal encoders \cite{lester2021power, zhou2022learning, jia2022visual}, thus failing to effectively integrate new information from the other modality during downstream transfer of dual-modal VLPMs. Recognizing these limitations, some researchers have proposed PT methods capable of tuning both modalities \cite{xing2023dual, chowdhury2023apollo, li2023efficient, khattak2023maple, zang2022unified}, including synchronous methods, such as MaPLe \cite{khattak2023maple} and UPT \cite{zang2022unified}. Synchronous PT methods aim to incorporate bimodal information into dual-modal prompts via a single set of learnable tokens, expecting that prompts from different modalities can remain semantically aligned on downstream tasks.

However, current dual-modal PT methods, whether synchronous or not, do not fit in well with fusion-based VLPMs, which benefit from a unique deep fusion structure that intricately aligns and couples features extracted from text and image encoders during pre-training \cite{li2022grounded, zhang2022glipv2, dou2022coarse}. Pre-trained on vast amounts of text-image pairs, the deep fusion structures learn dual-modal mappings and aligning space that closely approximates an ideal, unbiased estimator, thus offering powerful object retrieval capabilities. The aforementioned dual-modal PT methods require additional modal mapping substructures to remodel the aligning space for dual modalities, entailing the recoupling and redistribution of information from both modalities. We believe this strategy is suboptimal for fusion-based VLPMs. Firstly, it fails to fully utilize the pre-trained modal mapping inherent in deep fusion. Learning new modal mapping functions and aligning spaces for downstream tasks with limited data often results in approximate distributions that are inaccurate, leading to reduced transfer generalization and even poorer performance compared to single-modal PEFT methods. Secondly, additional modal mapping substructures incur extra training and storage costs. 

In this paper, we propose a novel prompt tuning method specially tailored for fusion-based VLPMs, called Synchronous Dual Prompt Tuning (SDPT). SDPT fine-tunes a single set of learnable tokens, named unified prototype tokens, to simultaneously acquire information from both text and image modalities on downstream tasks (See \cref{briefcompare}.(a)). To avoid the need for remodeling the aligning space, SDPT's unified prototype tokens are positioned in the fusion space, which is inherently suitable for aligning text and image information. Importantly, within the X-MHA operators of fusion-based VLPMs, there already exist pre-trained mappings from different modalities to the fusion space. Hence, we inherit these original pre-trained mappings in the fusion-based VLPMs to construct accurate inverse linear projections that necessitate no training for modal mapping. These projections synchronously remap and append the unified prototype tokens to text and image tokens, facilitating a shared downstream task semantics of text and image across different modal prompts. During tuning, the weights of the prototype tokens are updated to seamlessly integrate the dual modality information of new tasks. SDPT effectively preserves the unbiased modal mapping functions and aligning space established by fusion-based VLPMs for the first time, maximizing the use of pre-trained knowledge. Moreover, it reduces training costs by eliminating the need for learning new modal mapping substructures or modal aligning spaces, achieving better PEFT performance on new tasks for fusion-based VLPMs. The experimental results demonstrate that SDPT outperforms previous PEFT methods in assisting fusion-based VLPMs, including GLIP \cite{li2022grounded} and FIBER \cite{dou2022coarse}, to achieve optimal transfer effects with minimal trainable parameters across various experimental settings on COCO \cite{lin2014microsoft}, LVIS \cite{gupta2019lvis}, and ODinW13 \cite{li2022elevater} datasets (See \cref{briefcompare}.(b)). Furthermore, SDPT exhibits high generality and flexibility, showing compatibility with previously fine-tuned PEFT components, and can even be seamlessly integrated into standard self-training frameworks \cite{dopido2013semisupervised, wu2023zero}. Our contributions are summarized as follows: 
\begin{itemize}
    \item We propose a novel prompt tuning method, termed Synchronous Dual Prompt Tuning (SDPT), the first dual-modal PEFT approach specially designed for fusion-based VLPMs. SDPT achieves superior transfer performance than other PEFT methods with only 0.04\% of model parameters for training.
    \item SDPT constructs learnable unified prototype tokens within the deep fusion space for representing the aligned semantics of text and image. Furthermore, SDPT introduces accurate inverse linear projections that need no training based on established deep fusion.
    \item SDPT avoids learning inaccurate modal mapping or aligning distributions from limited downstream data, fully leveraging the pre-trained knowledge of fusion-based VLPMs. Extensive experiments have demonstrated the generality and flexibility of SDPT in various settings.
\end{itemize}

\section{Related Works}
Fusion-based visual-language pre-trained models (VLPMs), including GLIP \cite{li2022grounded}, GLIPv2 \cite{zhang2022glipv2}, GLIGEN \cite{li2023gligen}, and FIBER \cite{dou2022coarse}, are a specialized branch within VLPMs, focusing specifically on object-level vision tasks. Fusion-based VLPMs employ a grounded pre-training methodology involving both object detection and phrase grounding data. Furthermore, Fusion-based VLPMs' deep fusion, based on X-MHA layers, makes the learned vision representations specifically compatible with object-level tasks \cite{li2022grounded, dou2022coarse, li2023gligen}. Although Fusion-based VLPMs have demonstrated impressive zero-shot inference performance, there still is space for improvement in their transferability on downstream tasks. Currently, the efficient fine-tuning of Fusion-based VLPMs remains under-explored.

Large pre-trained models show outstanding abilities in NLP and CV, achieving impressive transfer performance \cite{devlin2018bert, radford2019language, brown2020language, wang2018glue, yang2019xlnet, zhou2022learning, jia2021scaling, carion2020end, dosovitskiy2010image, wang2021pyramid, xie2021segformer, yuan2021tokens, zheng2021rethinking}. However, fully fine-tuning these models for complex tasks is prohibitive \cite{he2021towards}. To address this, researchers have turned to parameter-efficient fine-tuning (PEFT) techniques, mainly divided into token-related prompt tuning (PT) and network-related methods. The former involves adding learnable tokens to input tokens on transformer architectures\cite{lester2021power, zhou2022learning, jia2022visual, xing2023dual, li2021prefix}, while the latter integrates small learnable structures or partially unfreezes weights for feature adjustment \cite{chen2022adaptformer, houlsby2019parameter, hu2021lora, gao2023clip}. Despite their effectiveness, these methods primarily suit single-branch architectures and face challenges in integrating new information from another modality in the context of dual-modal VLPMs \cite{zhou2022learning, jia2021scaling}. 

Some network-related methods like LoRA \cite{hu2021lora} and BitFit \cite{zaken2021bitfit} have good versatility and can be applied to dual-modal VLPMs, but their tuning performance still needs to be improved. Therefore, researchers have shifted focus to VLPM-specific dual-modal PT methods. Among these, some have proposed asynchronous approaches, adding learnable structures to both modal branches separately for dual-modal knowledge learning, such as DPT \cite{xing2023dual}, APOLLO \cite{chowdhury2023apollo}, and PMF \cite{li2023efficient}. However, these asynchronous methods require additional training resources and face challenges in semantically aligning bimodal information for downstream tasks through training. Thus, synchronous token-related PEFT methods like MaPLe \cite{khattak2023maple} and UPT \cite{zang2022unified} have been proposed to process bimodal information with a single set of learnable tokens, expecting that prompts from different modalities can be semantically aligned and meanwhile lowering the parameter count. Nonetheless, existing dual-modal PEFT approaches, synchronous or not, are not designed for fusion-based VLPMs, necessitating extra modal mapping substructures to recouple and redistribute the dual-modal information. This strategy often yields inaccurate dual-modal mapping or aligning distributions for fusion-based VLPMs because of the limited tuning data, leading to reduced transfer generalization and potentially degrading the performance. Additionally, it also increase the training costs. This paper proposes SDPT, a novel PT method tailored for fusion-based VLPMs, enhancing both training efficiency and transfer results over existing approaches.

\section{Method}
This section begins with a preliminary of key notations in fusion-based VLPMs, before introducing our SDPT.
\subsection{Preliminary}


\textbf{Fusion-based VLPMs.} In this paper, we use the GLIP \cite{li2022grounded} models to illustrate Fusion-based VLPMs \cite{li2022grounded,zhang2022glipv2,dou2022coarse,li2023gligen}. The image encoder of GLIP generates visual embeddings for different regions in the image to align with object words present in the text. Denoting text tokens and visual tokens as $P \in \mathbb{R}^{n \times d_{T}}$ and $R \in \mathbb{R}^{m \times d_{I}}$, where $n$ and $m$ represent the length of text tokens and image tokens respectively, and $d_{T}$, $d_{I}$ represents the dimension of the text and image latent space. GLIP's text and image encoder layers, respectively denoted as $ {\operatorname{enc}}_{T}^i\left(.\right)$ and ${\operatorname{enc}}_{I}^i\left(.\right)$, progressively encode $P$ and $R$. Setting the total number of GLIP's encoder layers as $L$, the deep fusion process of GLIP can be represented as:
\begin{equation}
\begin{aligned}
&P_{I2T}^i, R_{T2I}^i={\operatorname{X-MHA}}^{i+1}\left(P^i,R^i\right),\\
&P^{i+1},R^{i+1}=\operatorname{enc}_{T}^{i+1}\left(P^{i}+P_{I2T}^i\right), \operatorname{enc}_{I}^{i+1}\left(R^i+R_{T2I}^i\right),
\end{aligned}
\end{equation}
where $i \in \{0,1,\ldots,L-1\}$, $P^0$ and $R^0$ denote the tokens outputted from the pre-trained BERT \cite{devlin2018bert} and the pre-trained swin-transformer \cite{liu2021swin}, respectively. $\operatorname{X-MHA}^{i+1}$ represents bimodal cross multi-head attention in ${(i+1)}^{th}$ layer, which is specifically formularized as:
\begin{equation}
\begin{aligned}
   & P^{(q)}=P W^{(q, T)}, R^{(q)}=R W^{(q, I)}, \operatorname{Attn}=\frac{R^{(q)}\left(P^{(q)}\right)^{\top}}{\sqrt{d}},\\
    &R_{T2I}=\operatorname{Softmax}(\operatorname{Attn})P^{(v)}, P_{I2T}=\operatorname{Softmax}(\operatorname{Attn}^{\top})R^{(v)},
\end{aligned}
\end{equation}
where $\left\{W^{(q, \bmod)}:\bmod \in\{T, I\}\right\}$ represents the query transform matrices. After pre-training, both $P$ and $R$ are mapped from their individual modality spaces to a shared fusion space with a latent dimension of $d$, denoted as $P^{(q)}$ and $R^{(q)}$. $P^{(q)}$ and $R^{(q)}$ mutually serve as key vectors to compute attention weights of the coupling between the two modalities. Together with value matrices $P^{(v)}$ and $R^{(v)}$, the coupled vectors $P_{I2T}$ and $R_{T2I}$ are obtained. For simplicity, we have omitted noncritical transform matrices, like the value transform matrices, and all bias terms of the transform. This deep fusion process enables fusion-based VLPMs to achieve multi-level coupled modeling in the dual modalities of text and image, yielding superior object-level semantic features \cite{li2022grounded, zhang2022glipv2, dou2022coarse}.

\subsection{Synchronous Dual Prompt Tuning}

\begin{figure*}[t]
    \centering
    \includegraphics[width=\columnwidth]{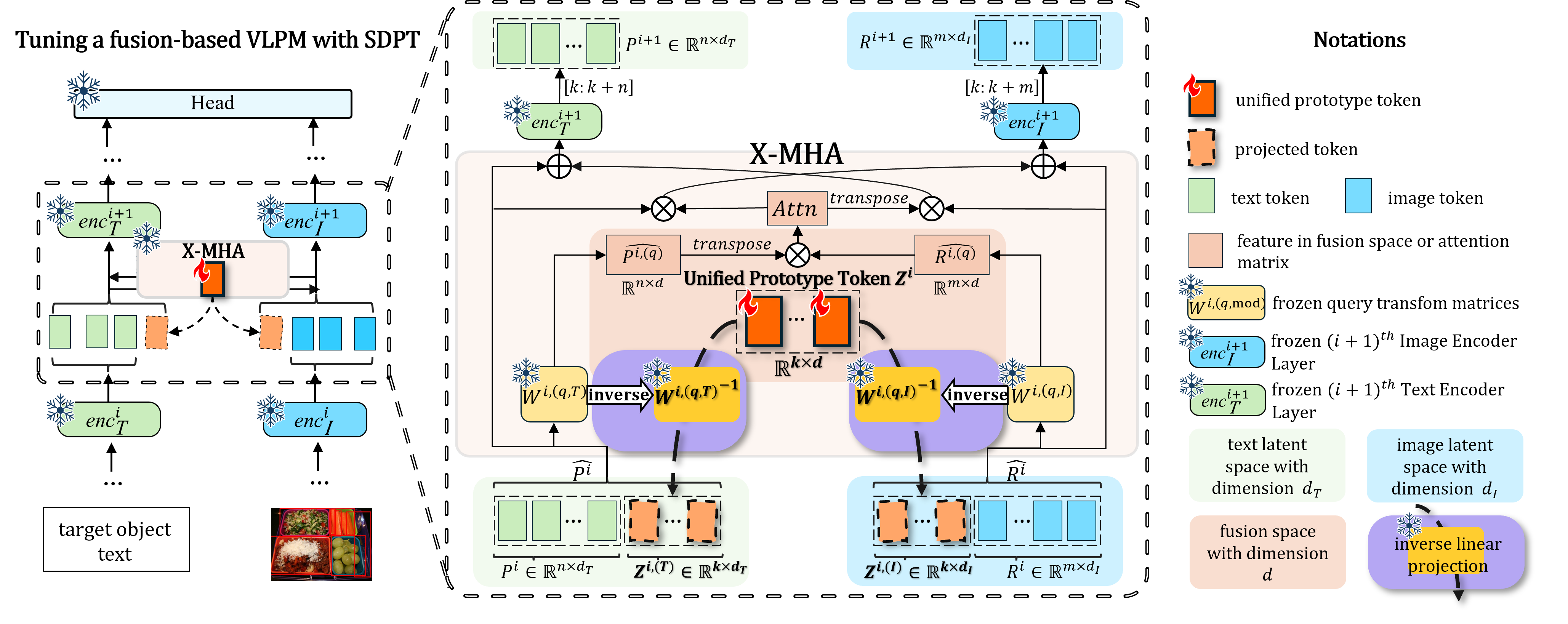}
    \caption{Detailed illustration of Synchronous Dual Prompt Tuning (SDPT). $\operatorname{X-MHA}$ refers to the cross attention layer. Unified prototype tokens $Z^i$ in each X-MHA layer are tuned through inverse linear projections to synchronously incorporate dual-modality knowledge for the new task while keeping the other parameters of the network frozen. }
    \label{fig:detail}
\end{figure*}

Existing dual-modal PT methods utilize additional modal mapping substructures to remodel the modal aligning space with limited data, which may lead to inaccurate mapping or aligning distribution and weaken PEFT effectiveness. To overcome these problems, we propose Synchronous Dual Prompt Tuning (SDPT) specifically designed for fusion-based VLPMs, detailed in \cref{fig:detail}.

\subsubsection{Unified Prototype Token} To reduce the number of trainable parameters and simplify the search for optimal hyperparameter configurations, we use a single set of modality-sharing learnable tokens. These are referred to as unified prototype tokens and are used for integrating dual-modality information synchronously, similar to existing methods \cite{khattak2023maple,zang2022unified}. In a GLIP model, text tokens $P^i\in \mathbb{R}^{n\times d_{T}}$ and image tokens $R^i\in \mathbb{R}^{m\times d_{I}}$ inherently reside in different latent spaces due to different encoding architectures. In practical downstream tasks, target category texts and images are expected to achieve semantic alignment within a specific latent space, namely the modal aligning space, which is assumed to contain latent vectors that effectively express the task. Previous approaches \cite{khattak2023maple,zang2022unified} necessitate unnecessary remodeling of this modal aligning space, which may impair PEFT performance. To circumvent this, we position the unified prototype tokens within the pre-established cross-attention space, i.e. the fusion space. This is feasible because the fusion space spanned by the X-MHA layer of the GLIP model with a dimension of $d$ is precisely the space where semantic alignment of $P^i$ and $R^i$ is achieved. Therefore, for the ${(i+1)}^{th}$ cross-attention layer $\operatorname{X-MHA}^{i+1}$, we introduce $k$ unified prototype tokens $Z^i\in \mathbb{R}^{k\times d}$, to sufficiently represent the information from both text and image modality in downstream tasks.

\subsubsection{Inverse Linear Projections} To enable the unified prototype tokens $Z^i$ to simultaneously acquire new knowledge from both the text ($T$) and image ($I$) modalities during end-to-end training, a transformation method is required to respectively map them into the latent spaces of $P^i$ and $R^i$ for the forward and backward computations. An intuitive strategy involves using two learnable modal mapping substructures to transform $Z^i$ from the $d$-dimensional fusion space to the latent spaces with $d_T$ and $d_I$, respectively, as seen in existing synchronous dual-modal methods \cite{khattak2023maple,zang2022unified}. However, this strategy suffers from learning biased mapping distributions with highly limited downstream data, which may harm the transfer generalization of fusion-based VLPMs. In fact, the target mappings are inherently embedded in the query transform matrices $\left\{W^{(q, \bmod)}:\bmod \in\{T, I\}\right\}$ in X-MHA established by GLIP during pre-training. The query transformations from $P$ and $R$ to $P^{(q)}$ and $R^{(q)}$ in the X-MHA are linear. Observing this, we establish two inverse linear projections based on $\left\{W^{(q, \bmod)}:\bmod \in\{T, I\}\right\}$, enabling synchronous mapping of $Z^i$ back to the text and image latent spaces without introducing additional trainable parameters. Omitting the layer notation $i$, $Z^i$ is simply denoted as $Z$, this process can be formalized as follows:
\begin{equation}
\begin{aligned}
&W^{(q, T)^{-1}}, W^{(q, I)^{-1}}=\operatorname{Pinv}(W^{(q, T)}),\operatorname{Pinv}(W^{(q, I)}),\\
&Z^{(T)},Z^{(I)}=(Z-B^{(q, T)}) W^{(q, T)^{-1}}, (Z-B^{(q, I)}) W^{(q, I)^{-1}}.
\end{aligned}
\end{equation}
The pseudo-inverse operation $\operatorname{Pinv}(\cdot)$ is applied once at the beginning of the tuning process to obtain $\left\{W^{(q, \bmod)^{-1}}:\bmod \in\{T, I\}\right\}$. $\{B^{(q,\bmod)}$: $\bmod \in\{T, I\}\}$ is the bias of the pre-trained query transform. This transformation aligns $Z^{i,(T)} \in \mathbb{R}^{k \times d_{T}}$ with text tokens $P^i$, as well as $Z^{i,(I)} \in \mathbb{R}^{k \times d_{I}}$ with image tokens $R^i$, in the same latent space. By concatenating $Z^{i,(T)}$ and $ Z^{i,(I)}$ with $P^i$ and $R^i$ respectively, we form $\hat{P^i}$=$\left[ Z^{i,(T)},P^i \right] \in \mathbb{R}^{(k+n) \times d_{T}}$ and $\hat{R^i}$=$\left[ Z^{i,(I)},R^i \right] \in \mathbb{R}^{(k+m) \times d_{I}}$, enabling joint forward computation. Since $P^i$ and $R^i$ vary, forming $\hat{P^i}$ and $\hat{R^i}$ from the unified prototype token $Z^i$ maintains “one-to-multi” mappings, avoiding undesirable bias. This process can be formalized as follows:
\begin{equation}
\begin{aligned}
&\hat{P_{I2T}^i}, \hat{R_{T2I}^i}={\operatorname{X-MHA}}^{i+1}\left(\hat{P^i},\hat{R^i}\right),\\
&P^{i+1}=\operatorname{enc}_{T}^{i+1}\left(\hat{P^i}+\hat{P_{I2T}^i}\right)[k:k+n],\\&R^{i+1}=\operatorname{enc}_{I}^{i+1}\left(\hat{R^i}+\hat{R_{T2I}^i}\right)[k:k+m],\\
&y'= \operatorname{Head}(P^L,R^L).
\end{aligned}
\end{equation}
$[:]$ denotes slicing operation. $P^{i+1} \in \mathbb{R}^{n \times d_{T}}$, $R^{i+1} \in \mathbb{R}^{m \times d_{I}}$ are then processed through subsequent layers, then ultimately inputted into the $\operatorname{Head(\cdot)}$ layer to generate $y'$. The loss between $y'$ and the ground truth $y$ is computed, updating $Z^i$'s weights to synchronously incorporate dual-modal knowledge for the new task while keeping the network's other parameters frozen. Compared to learning new modal mapping functions, the aforementioned inverse linear projections directly employ the mappings established on pre-training data, which closely approximates the ideal unbiased distribution. This approach fully capitalizes on GLIP's transfer generalization capabilities. Additionally, these inverse linear projections need no training, further reducing the number of trainable parameters.
\\

\noindent SDPT tunes input features from both modalities and adheres more closely to the characteristics of fusion-based VLPMs and their transfer requirements. On the one hand, the unified prototype tokens $Z$ within the established fusion space and the inverse linear projections preserve the pre-trained text-image aligning knowledge. On the other hand, the two components further reduce the tuning difficulty and the training costs,  enhancing tuning efficiency with fewer parameters on fusion-based VLPMs. Notably, SDPT exhibits impressive fine-tuning performance even when the length of unified prototype tokens $k$ is small or $Z^i$ is incorporated into just a selection of X-MHA layers, achieving the best trade-off between performance and model training cost. SDPT is compatible with any fusion-based VLPMs featuring X-MHA, including FIBER \cite{dou2022coarse}.

\section{Experiments}

\subsection{Downstream Tasks}
We primarily validated the superiority of our proposed SDPT in fusion-based VLPMs on three datasets, namely MS COCO 2017 \cite{lin2014microsoft}, LVIS \cite{gupta2019lvis}, and ODinW13 \cite{li2022elevater}. Details are provided in the supplementary materials.

\subsection{Comparison methods}
We evaluated SDPT against the baseline, linear probing and full fine-tuning, and the representative PEFT methods, categorizing them by the modality they tune --- distinguishing between dual-modal and single-branch (text or image) PEFT approaches. For text modality, we chose the token-related CoOp \cite{zhou2022learning} and the network-related adapter \cite{houlsby2019parameter}. Similarly, VPT \cite{jia2022visual} and Adaptformer \cite{chen2022adaptformer} were selected for image modality.  Regarding dual-modal PEFT methods, we applied versatile network-related methods LoRA \cite{hu2021lora} and BitFit \cite{zaken2021bitfit} to fusion-based VLPMs' dual-modal encoders for comparison, and assessed their performance on the cross-attention layers, denoted as "LoRA$_{XMHA}$" and "BitFit$_{XMHA}$". We also compared SDPT with dual-modal PEFT methods designed for standard VLPMs, including asynchronous methods DPT \cite{xing2023dual}, APOLLO \cite{chowdhury2023apollo}, PMF \cite{li2023efficient}, and synchronous methods MaPLe \cite{khattak2023maple} and UPT \cite{zang2022unified}. Additionally, we also trivially combined single-modal methods of different modalities to explore the effects of such straightforward combinations for dual-modal fusion-based VLPMs, with these results detailed in the supplementary materials.

\subsubsection{Implementation}
Our main experiments utilized the pre-trained GLIP-L model, using evaluation metrics from the original GLIP study \cite{li2022grounded} plus inference FLOPs and trainable parameter volume for efficiency. Experimental settings, unless specialized, involved incorporating unified prototype tokens of length $k=120$ into each X-MHA layer of GLIP-L, which was obtained through cross-validation on COCO. We used uniform random initialization between -1 and 1 for the unified prototype tokens. More details, including hyperparameters for comparison methods, are provided in the supplementary materials.

\begin{table*}[t]
  \caption{Comparison with other PEFT methods. Note: results of linear probing on ODinW13 (line 1) and full fine-tuning on COCO and ODinW13 (line 6) are from the original GLIP paper \cite{li2022grounded}, with "\textbackslash{}" indicating data not reported. "\#Par" denotes the number of trainable parameters, with values in "($\cdot$)" representing the proportion of trainable parameters relative to the total model parameters. The values in the ODinW13 column represent the average mAP across 13 subtasks. Our method, SDPT, is highlighted with a gray background. The best and second-best values are highlighted in \textbf{bold} and {\ul underlined}, respectively.}
  \label{fullcomparison}
  \centering
  \resizebox{\textwidth}{!}{
  \renewcommand{\arraystretch}{1}
\begin{tabular}{cc|cc|cccc|cccc|ccccc}
\hline
\multicolumn{2}{c|}{} &  &  & \multicolumn{4}{c|}{\textbf{COCO}} & \multicolumn{4}{c|}{\textbf{LVIS}} & \multicolumn{5}{c}{\textbf{ODinW13}} \\ \cline{5-17} 
\multicolumn{2}{c|}{\multirow{-2}{*}{\textbf{Method}}} & \multirow{-2}{*}{\scriptsize{\textbf{FLOPs(G)}}} & \multirow{-2}{*}{\scriptsize{\textbf{\#Par(M)}}} & \textbf{mAP} & \textbf{AP50} & \textbf{AP75} & \textbf{AR} & \textbf{AP} & \textbf{APr} & \textbf{APc} & \textbf{APf} & \textbf{\scriptsize{1-shot}} & \textbf{\scriptsize{3-shot}} & \textbf{\scriptsize{5-shot}} & \textbf{\scriptsize{10-shot}} & \textbf{\scriptsize{full-shot}} \\ \hline
\multicolumn{2}{c|}{\textbf{Linear probing}} & \textbf{724.30} & 1.96\tiny{(0.49\%)} & 52.7 & 69.9 & 58.1 & 69.4 & 33.8 & 25.3 & 29.8 & 40.6 & 54.1 & 54.7 & 55.0 & 55.8 & 59.2 \\ \hline
\multicolumn{1}{c|}{} & \textbf{Adaptformer\cite{chen2022adaptformer}} & 780.87 & 56.60\tiny{(14.24\%)} & 55.8 & 73.4 & 60.9 & 71.4 & 39.4 & 29.8 & 34.7 & 43.8 & 47.3 & 55.0 & 56.1 & 60.1 & 68.0 \\
\multicolumn{1}{c|}{\multirow{-2}{*}{\textbf{Image}}} & \textbf{VPT\cite{jia2022visual}} & 1317.41 & 4.26\tiny{(1.07\%)} & 57.4 & 75.0 & 62.5 & 72.5 & 40.7 & 29.7 & 32.3 & 44.2 & 58.9 & 61.3 & 63.6 & 65.2 & 68.8 \\ \hline
\multicolumn{1}{c|}{} & \textbf{Adapter\cite{houlsby2019parameter}} & {\ul 724.31} & 10.62\tiny{(2.67\%)} & 57.5 & 75.3 & 62.9 & 72.4 & 40.4 & 29.8 & 34.3 & 43.9 & 59.8 & 61.6 & 62.5 & 65.0 & 69.2 \\
\multicolumn{1}{c|}{\multirow{-2}{*}{\textbf{Text}}} & \textbf{CoOp\cite{zhou2022learning}} & 1306.24 & 0.52\tiny{(0.13\%)} & 55.4 & 73.2 & 60.7 & 71.1 & 39.5 & 29.8 & 33.4 & 43.1 & 59.6 & 61.6 & 63.1 & 65.5 & 68.8 \\ \hline
\multicolumn{1}{c|}{} & \textbf{Full FT} & \textbf{724.30} & 397.59\tiny{(100\%)} & \textbf{60.8} & \textbackslash{} & \textbackslash{} & \textbackslash{} & {\ul 41.2} & 31.2 & 34.2 & {\ul 45.0} & 59.9 & 62.1 & 64.2 & 64.9 & 68.9 \\
\multicolumn{1}{c|}{} & \textbf{LoRA\cite{hu2021lora}} & \textbf{724.30} & 2.03\tiny{(0.51\%)} & 56.5 & 74.1 & 61.8 & 71.0 & 40.8 & 30.4 & 33.7 & 43.5 & 60.9 & 61.7 & 63.1 & 64.8 & 66.9 \\
\multicolumn{1}{c|}{} & \textbf{BitFit\cite{zaken2021bitfit}} & \textbf{724.30} & 0.42\tiny{(0.11\%)} & 54.5 & 73.0 & 60.1 & 70.6 & 39.1 & 29.7 & 34.2 & 43.1 & 61.2 & 61.7 & 64.1 & {\ul 65.6} & 68.3 \\
\multicolumn{1}{c|}{} & \textbf{LoRA$_{XMHA}$\cite{hu2021lora}} & \textbf{724.30} & 1.58\tiny{(0.40\%)} & 55.4 & 72.8 & 60.5 & 71.2 & 38.6 & 29.8 & 35.1 & 43.4 & 60.0 & 61.3 & 63.2 & 63.3 & 68.6 \\
\multicolumn{1}{c|}{} & \textbf{BitFit$_{XMHA}$\cite{zaken2021bitfit}} & \textbf{724.30} & 0.21\tiny{(0.05\%)} & 52.6 & 69.5 & 57.3 & 70.3 & 36.1 & 26.7 & 29.7 & 41.6 & {\ul 61.5} & 62.1 & 62.9 & 63.5 & 65.2 \\
\multicolumn{1}{c|}{} & \textbf{DPT\cite{xing2023dual}} & 925.17 & 80.71\tiny{(20.30\%)} & 51.1 & 67.9 & 57.1 & 67.3 & 35.3 & 26.1 & 30.3 & 41.5 & 59.0 & 59.7 & 60.6 & 61.6 & 65.2 \\
\multicolumn{1}{c|}{} & \textbf{Apollo\cite{chowdhury2023apollo}} & 1150.92 & 3.18\tiny{(0.80\%)} & 54.3 & 73.1 & 58.5 & 69.5 & 37.4 & {\ul 31.5} & 34.0 & 41.4 & 55.8 & 60.2 & 60.4 & 62.5 & 66.7 \\
\multicolumn{1}{c|}{} & \textbf{PMF\cite{li2023efficient}} & 730.59 & 1.53\tiny{(0.38\%)} & 56.2 & 74.2 & 60.2 & 70.4 & 39.6 & 30.8 & 33.2 & 42.2 & 58.6 & 61.1 & 61.8 & 64.0 & 67.4 \\
\multicolumn{1}{c|}{} & \textbf{UPT\cite{zang2022unified}} & 785.32 & 2.75\tiny{(0.69\%)} & 56.8 & 74.5 & 61.3 & 71.7 & 40.2 & 29.8 & 33.4 & 43.3 & 59.6 & 61.8 & 63.3 & 63.9 & 67.7 \\
\multicolumn{1}{c|}{} & \textbf{MaPLe\cite{khattak2023maple}} & 801.41 & 2.96\tiny{(0.74\%)} & 57.2 & 75.1 & 62.2 & 72.3 & 40.8 & 30.6 & 33.8 & 43.7 & 60.8 & 61.9 & 62.8 & 64.1 & 68.7 \\
\multicolumn{1}{c|}{} & \cellcolor[HTML]{D9D9D9}\textbf{SDPT:k=10} & \cellcolor[HTML]{D9D9D9}724.77 & \cellcolor[HTML]{D9D9D9}\textbf{0.16\tiny{(0.04\%)}} & \cellcolor[HTML]{D9D9D9}57.6 & \cellcolor[HTML]{D9D9D9}{\ul 75.4} & \cellcolor[HTML]{D9D9D9}{\ul 63.0} & \cellcolor[HTML]{D9D9D9}{\ul 72.8} & \cellcolor[HTML]{D9D9D9}{\ul 41.2} & \cellcolor[HTML]{D9D9D9}{\ul 31.5} & \cellcolor[HTML]{D9D9D9}{\ul 34.9} & \cellcolor[HTML]{D9D9D9}{\ul 45.0} & \cellcolor[HTML]{D9D9D9}{\ul 61.5} & \cellcolor[HTML]{D9D9D9}{\ul 62.2} & \cellcolor[HTML]{D9D9D9}{\ul 64.3} & \cellcolor[HTML]{D9D9D9}{\ul 65.6} & \cellcolor[HTML]{D9D9D9}{\ul 69.5} \\
\multicolumn{1}{c|}{\multirow{-12}{*}{\textbf{Dual}}} & \cellcolor[HTML]{D9D9D9}\textbf{SDPT:k=120} & \cellcolor[HTML]{D9D9D9}738.36 & \cellcolor[HTML]{D9D9D9}1.97\tiny{(0.50\%)} & \cellcolor[HTML]{D9D9D9}{\ul 58.0} & \cellcolor[HTML]{D9D9D9}\textbf{75.8} & \cellcolor[HTML]{D9D9D9}\textbf{63.2} & \cellcolor[HTML]{D9D9D9}\textbf{73.1} & \cellcolor[HTML]{D9D9D9}\textbf{41.4} & \cellcolor[HTML]{D9D9D9}\textbf{31.8} & \cellcolor[HTML]{D9D9D9}\textbf{35.2} & \cellcolor[HTML]{D9D9D9}\textbf{45.1} & \cellcolor[HTML]{D9D9D9}\textbf{61.6} & \cellcolor[HTML]{D9D9D9}\textbf{64.4} & \cellcolor[HTML]{D9D9D9}\textbf{64.4} & \cellcolor[HTML]{D9D9D9}\textbf{66.4} & \cellcolor[HTML]{D9D9D9}\textbf{71.2} \\ \hline
\end{tabular}
}
\end{table*}

\begin{figure}[ht]
    \centering
    \includegraphics[width=0.98\columnwidth]{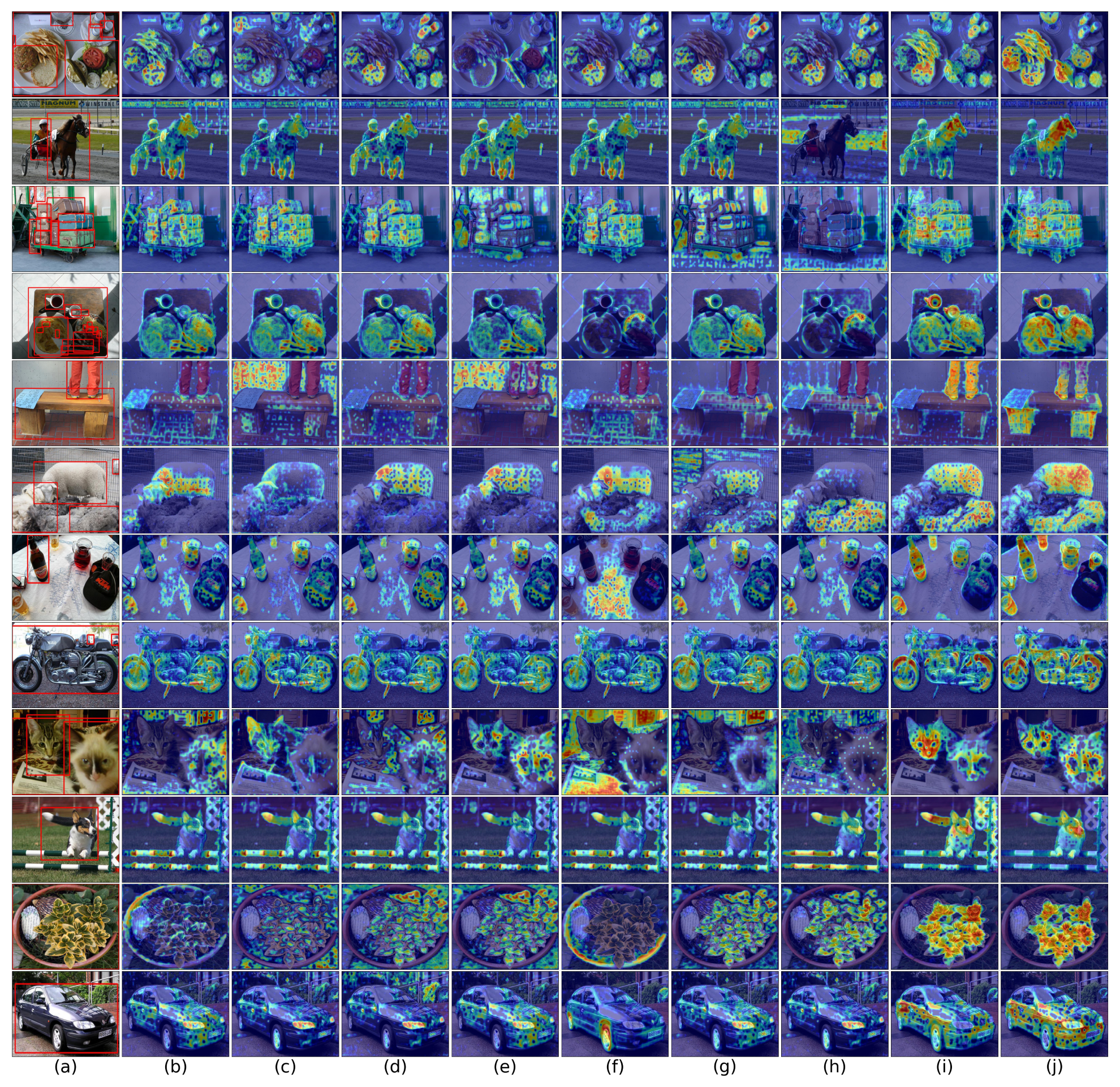}
    \caption{Comparison of attention map visualization of the different methods on LVIS (line 1\textasciitilde 6) and PascalVOC (line 7\textasciitilde 12). (a) Original image and ground truths, (b) LoRA, (c) BitFit, (d) DPT, (e) Apollo, (f) PMF, (g) UPT, (h) MaPLe, (i) SDPT (k=10), (j) SDPT (k=120 for LVIS, 200 for PascalVOC). Ground truths are marked by red boxes.} 
    \label{fig:cam}
\end{figure}

\subsubsection{Main Comparison Results}

The comparative results between SDPT and other PEFT approaches are presented in \cref{fullcomparison}, showcasing full-shot transfer tuning across three benchmarks and few-shot experiments on ODinW13 as per the GLIP paper \cite{li2022grounded}. To be noted that: (1) The outcomes of linear probing on ODinW13 (line 1) and full fine-tuning on COCO and ODinW13 (line 6) are from the GLIP study, with "\textbackslash{}" marking missing data. (2) Due to ODinW13's task complexity, an aggressive cross-validation strategy was employed on one of its sub-datasets, Aquarium, to determine the hyperparameter $k$ (see \cref{sec:token length}). The superior results on ODinW13, listed in the final row of the table, were achieved with $k=200$, whereas COCO and LVIS employed $k=120$. (3) The inference FLOPs and the number of trainable parameters for the final row are reported under $k=120$. (4) Since LoRA and BitFit do not alter the model's parameter count, their inference FLOPs are consistent with those of linear probing and full fine-tuning.

The table results reveal several noteworthy phenomena: (1) MaPLe and UPT achieve superior PEFT performance compared to asynchronous dual-modal methods, demonstrating the advantage of synchronous dual-modal information integration. (2) However, methods specifically designed for dual-modal VLPMs, including synchronous MaPLe and UPT, sometimes do not outperform single-modal methods on fusion-based VLPMs. This occurs because they approximate a biased distribution, inadvertently reducing the generalization ability. In contrast, single-modal PEFT methods, along with LoRA and BitFit, achieve higher performance because they do not remodel the aligning space. (3) "LoRA$_{XMHA}$" and "BitFit$_{XMHA}$" show decreased effectiveness compared to their original versions. This phenomenon indicates that tuning the fusion space can reduce fusion-based VLPMs' generalization and transfer performance.

SDPT, by situating unified prototype tokens within the established fusion space and constructing the inverse linear projections from pre-trained X-MHA, achieves optimal results with the minimal number of trainable parameters required. Notably excelling in LVIS and ODinW13, SDPT surpasses traditional fine-tuning, marking a PEFT breakthrough for fusion-based VLPMs. Remarkably, even with $k=10$ (0.04\% of full fine-tuning parameters), SDPT achieves significant tuning effects. Furthermore, SDPT delivers optimal performance across few-shot settings, especially showing substantial improvement over the baseline in the one-shot scenario. This underscores SDPT's capability to utilize pre-trained knowledge for superior transfer results with limited data. We illustrate SDPT's effectiveness through attention heatmaps for category text on the GLIP-L model after fine-tuning on LVIS or PascalVOC (one subset of ODinW13), shown in \cref{fig:cam}. Unlike other dual-modal methods that show dispersed attention, indicating learning inaccurate distribution, SDPT focuses attention more precisely on target objects. Detailed few-shot comparisons and more visualizations are available in the supplementary materials.



\begin{table}[t]
\centering 
\begin{minipage}[b]{0.48\linewidth} 
\centering 
\caption{Performance on different fusion-based VLPMs. Note: results of "Zero-shot" (line 1) and "Full fine-tuning" (line 2) are from the original papers \cite{li2022grounded,dou2022coarse}. SDPT is highlighted with a gray background. The best and second-best values are highlighted in \textbf{bold} and {\ul underlined}, respectively.}
\label{tab:backbone}
\resizebox{\columnwidth}{!}{
\renewcommand\arraystretch{1}
\begin{tabular}{c|cccc}
\hline
 & \multicolumn{4}{c}{\textbf{Model (mAP on COCO)}} \\ \cline{2-5} 
\multirow{-2}{*}{\textbf{Methods}} & \scriptsize{\textbf{Glip-T(A)}} & \scriptsize{\textbf{GLIP-T}} & \scriptsize{\textbf{GLIP-L}} & \scriptsize{\textbf{FIBER-B}} \\ \hline
\textbf{Zero-shot} & 42.9 & 46.6 & 49.8 & 49.3 \\
\textbf{Full FT} & {\ul 52.9} & {\ul 55.2} & \textbf{60.8} & \textbf{58.4} \\
\textbf{Adaptformer\cite{chen2022adaptformer}} & 50.4 & 53.6 & 55.8 & 56.1 \\
\textbf{CoOp\cite{zhou2022learning}} & 48.4 & 52.2 & 55.4 & 55.8 \\
\textbf{LoRA\cite{hu2021lora}} & 50.1 & 53.8 & 56.5 & 55.2 \\
\textbf{BitFit\cite{zaken2021bitfit}} & 48.3 & 51.9 & 54.5 & 53.7 \\
\textbf{LoRA$_{XMHA}$\cite{hu2021lora}} & 47.4 & 52.2 & 55.4 & 53.1 \\
\textbf{BitFit$_{XMHA}$\cite{zaken2021bitfit}} & 43.7 & 47.9 & 52.6 & 51.0 \\
\textbf{DPT\cite{xing2023dual}} & 47.5 & 50.1 & 51.1 & 51.3 \\
\textbf{Apollo\cite{chowdhury2023apollo}} & 46.7 & 50.4 & 54.7 & 54.3 \\
\textbf{PMF\cite{li2023efficient}} & 49.8 & 52.6 & 56.6 & 55.3 \\
\textbf{UPT\cite{zang2022unified}} & 48.8 & 52.3 & 56.8 & 53.6 \\
\textbf{MaPLe\cite{khattak2023maple}} & 48.5 & 51.8 & 57.2 & 55.4 \\
\rowcolor[HTML]{D9D9D9} 
\textbf{SDPT} & \textbf{53.2} & \textbf{55.6} & {\ul 58.0} & {\ul 57.1} \\ \hline
\end{tabular}
}
\end{minipage}
\hfill 
\begin{minipage}[b]{0.49\linewidth} 
\centering 
\caption{Performance on self-training settings on Aquarium. Note: results of "Full fine-tuning" (line 6) are from the original papers \cite{li2022grounded}, with "\textbackslash{}" indicating data not reported. SDPT is highlighted with a gray background. The best and second-best values are highlighted in \textbf{bold} and {\ul underlined}, respectively.}
\label{tab:self-training}
\resizebox{0.825\columnwidth}{!}{
\renewcommand\arraystretch{1}
\begin{tabular}{cc|cccc}
\hline
\multicolumn{2}{c|}{} & \multicolumn{4}{c}{\textbf{Aquarium}} \\ \cline{3-6} 
\multicolumn{2}{c|}{\multirow{-2}{*}{\textbf{Method}}} & \textbf{mAP} & \textbf{AP50} & \textbf{AP75} & \textbf{AR} \\ \hline
\multicolumn{2}{c|}{\textbf{Linear probing}} & 43.7 & 68.2 & 48.5 & 51.3 \\ \hline
\multicolumn{1}{c|}{} & \textbf{Adaptformer\cite{chen2022adaptformer}} & 53.5 & 78.9 & 58.1 & 67.9 \\
\multicolumn{1}{c|}{\multirow{-2}{*}{\textbf{Image}}} & \textbf{VPT\cite{jia2022visual}} & 55.2 & 80.1 & 57.3 & 68.0 \\ \hline
\multicolumn{1}{c|}{} & \textbf{Adapter\cite{houlsby2019parameter}} & 54.2 & 78.4 & 56.1 & 67.7 \\
\multicolumn{1}{c|}{\multirow{-2}{*}{\textbf{Text}}} & \textbf{CoOp\cite{zhou2022learning}} & 55.1 & 78.6 & 58.0 & 68.0 \\ \hline
\multicolumn{1}{c|}{} & \textbf{Full FT} & 56.6 & \textbackslash{} & \textbackslash{} & \textbackslash{} \\
\multicolumn{1}{c|}{} & \textbf{LoRA\cite{hu2021lora}} & 55.8 & 81.4 & 60.5 & 69.0 \\
\multicolumn{1}{c|}{} & \textbf{BitFit\cite{zaken2021bitfit}} & 54.3 & 80.9 & 56.3 & 68.9 \\
\multicolumn{1}{c|}{} & \textbf{LoRA$_{XMHA}$\cite{hu2021lora}} & 56.6 & 83.4 & 59.9 & 67.9 \\
\multicolumn{1}{c|}{} & \textbf{BitFit$_{XMHA}$\cite{zaken2021bitfit}} & 50.2 & 75.6 & 51.8 & 68.0 \\
\multicolumn{1}{c|}{} & \textbf{DPT\cite{xing2023dual}} & 51.2 & 77.1 & 53.6 & 66.9 \\
\multicolumn{1}{c|}{} & \textbf{Apollo\cite{chowdhury2023apollo}} & 51.6 & 78.1 & 54.5 & 67.5 \\
\multicolumn{1}{c|}{} & \textbf{PMF\cite{li2023efficient}} & 53.9 & 81.4 & 59.6 & 68.4 \\
\multicolumn{1}{c|}{} & \textbf{UPT\cite{zang2022unified}} & 56.9 & 83.9 & \textbf{61.6} & 69.2 \\
\multicolumn{1}{c|}{} & \textbf{MaPLe\cite{khattak2023maple}} & {\ul 57.1} & {\ul 84.1} & 60.2 & {\ul 69.5} \\
\multicolumn{1}{c|}{\multirow{-11}{*}{\textbf{Dual}}} & \cellcolor[HTML]{D9D9D9}\textbf{SDPT} & \cellcolor[HTML]{D9D9D9}\textbf{57.2} & \cellcolor[HTML]{D9D9D9}\textbf{85.0} & \cellcolor[HTML]{D9D9D9}{\ul 61.1} & \cellcolor[HTML]{D9D9D9}\textbf{69.8} \\ \hline
\end{tabular}
}
\end{minipage}
\end{table}

\subsubsection{Generality and Flexibility} We validated the generality and flexibility of SDPT under various settings.
\begin{table}[t]
\caption{Compatibility with other fine-tuned PEFT modules, with the COCO \cite{lin2014microsoft} as an old task and the Aquarium as a new task. PEFT components with subscript "COCO" refer to those fine-tuned on the old task.}
\label{tab:compatibility}
\begin{center}
\resizebox{0.8\columnwidth}{!}{
\renewcommand\arraystretch{1}
\begin{tabular}{c|c|cc}
\toprule[1.2pt]
\multirow{2}{*}{\textbf{Method}} & \multirow{2}{*}{\textbf{Setting}} & \textbf{COCO} & \textbf{Aquarium} \\
 &  & \textbf{mAP} & \textbf{mAP} \\ \midrule[1.1pt]
\scriptsize{\textbf{GLIP\_L + Adapter$_{COCO}$}} & \begin{tabular}[c]{@{}c@{}}Adapter$_{COCO}$ has been   fine-tuned on \\      COCO. Zero-shot reference on Aquarium\end{tabular} & 56.8 & 30.7 \\ \hline
\scriptsize{\textbf{GLIP\_L + Adapter$_{COCO}$ + Adapter}} & \begin{tabular}[c]{@{}c@{}}Keep other parameters   frozen\\      and only tune a new Adapter on Aquarium\end{tabular} & 35.4 & 55.6 \\ \hline
\scriptsize{\textbf{GLIP\_L + Adapter$_{COCO}$ + SDPT}} & \begin{tabular}[c]{@{}c@{}}Keep other parameters   frozen\\      and only tune SDPT on Aquarium\end{tabular} & 53.2 & 58.7 \\ \midrule[1.1pt] 
\scriptsize{\textbf{GLIP\_L + VPT$_{COCO}$}} & \begin{tabular}[c]{@{}c@{}}VPT$_{COCO}$ has been fine-tuned   on \\      COCO. Zero-shot reference on Aquarium\end{tabular} & 55.4 & 27.6 \\ \hline
\scriptsize{\textbf{GLIP\_L + VPT$_{COCO}$ + VPT}} & \begin{tabular}[c]{@{}c@{}}Keep other parameters   frozen\\      and only tune a new VPT on Aquarium\end{tabular} & 30.8 & 53.1 \\ \hline
\scriptsize{\textbf{GLIP\_L + VPT$_{COCO}$ + SDPT}} & \begin{tabular}[c]{@{}c@{}}Keep other parameters   frozen\\      and only tune SDPT on Aquarium\end{tabular} & 51.9 & 58.3 \\ \bottomrule[1.2pt]
\end{tabular}
}
\end{center}
\end{table}

\textbf{Different fusion-based VLPMs.} We explored the full-shot tuning impacts of SDPT on various fusion-based VLPMs using the COCO benchmark, with results detailed in \cref{tab:backbone}. For FIBER \cite{dou2022coarse}, our approach involved incorporating the unified prototype tokens within the cross-attention modules of its dual-branch backbone, keeping other configurations unchanged. It is evident that SDPT consistently achieves optimal PEFT results across all fusion-based VLPMs, demonstrating its ability to effectively leverage the common characteristics of fusion-based VLPMs, showcasing its strong generality.

\textbf{Self-training setting.} SDPT also enhances the potential for zero-shot inference in fusion-based VLPMs. We conducted experiments on one of the ODinW13 subsets, Aquarium, to illustrate this point. After incorporating comparison PEFT methods into GLIP models, we employed the predictions generated by the original GLIP model as pseudo labels for self-training, as outlined in \cite{dopido2013semisupervised}. The experimental results are shown in \cref{tab:self-training}. Our SDPT, when incorporated into GLIP models for self-training, achieves remarkable unsupervised object detection performance, unleashing the zero-shot inference potential of fusion-based VLPMs. These findings further underscore SDPT's generality and flexibility. Implementation details on self-training settings are in the supplementary material.

\textbf{Compatibility with other fine-tuned PEFT methods.}
Our Synchronous Dual Prompt Tuning (SDPT) is compatible with other PEFT methods that have already been fine-tuned on an old task. Importantly, fine-tuning SDPT on a new task does not significantly diminish the model's performance on the old task. Experimental results are detailed in \cref{tab:compatibility}. The experiments were conducted on GLIP-L, using the COCO as the old task and the Aquarium as the new task. PEFT components with subscript "COCO" refer to those fine-tuned on the old task. The results show that our SDPT can seamlessly integrate with a pre-trained Adapter or VPT component. Training the unified prototype tokens while keeping the Adapter$_{COCO}$ or VPT$_{COCO}$ frozen, SDPT achieves satisfactory performance on both the old and the new tasks. In contrast, a new Adapter or VPT component fails to achieve a balanced performance between the old and new tasks. These outcomes underscore SDPT's remarkable flexibility and its capacity to preserve the generalization ability of fusion-based VLPMs. Furthermore, SDPT's compatibility with pre-trained PEFT components facilitates flexible deployment in practical applications.

\subsubsection{Ablation Studies}

\begin{table}[t]
\centering 
\begin{minipage}[b]{0.47\linewidth} 
\centering 
\caption{Effectiveness of synchronous dual-modal knowledge incorporation. "Sync." indicates whether the knowledge incorporating strategy is synchronous spatially and temporally. The best values are marked in \textbf{bold}.}
\label{tab:synchrounous}
\resizebox{\columnwidth}{!}{
\renewcommand\arraystretch{1}
\begin{tabular}{c|c|c|cccc}
\hline
\multirow{2}{*}{\textbf{Setting}} & \multirow{2}{*}{\textbf{Sync.}} & \multirow{2}{*}{\textbf{\scriptsize{\#Par(M)}}} & \multicolumn{4}{c}{\textbf{Aquarium}} \\
 &  &  & \textbf{mAP} & \textbf{AP50} & \textbf{AP75} & \textbf{AR} \\ \hline
\textbf{\begin{tabular}[c]{@{}c@{}}Unshared\\ prototype tokens\end{tabular}} & \textbf{×} & 6.56 & 53.3 & 80.2 & 56.3 & 66.1 \\ \hline
\textbf{\begin{tabular}[c]{@{}c@{}}Asynchronous\\ training\end{tabular}} & \textbf{×} & \textbf{3.28} & 54.8 & 81.7 & 57.4 & 67.8 \\ \hline
\textbf{\begin{tabular}[c]{@{}c@{}}Standard\\ setting\end{tabular}} & \checkmark & \textbf{3.28} & \textbf{57.7} & \textbf{85.0} & \textbf{60.6} & \textbf{70.5} \\ \hline
\end{tabular}
}
\end{minipage}
\hfill 
\begin{minipage}[b]{0.49\linewidth} 
\centering 
\caption{Effectiveness of inverse linear projections. We compared our inverse linear projection with the learnable linear projection. "Modal" indicates the modal branch receiving the projected tokens. The best values are marked in \textbf{bold}.}
\label{tab:tuningfree}
\resizebox{\columnwidth}{!}{
\renewcommand\arraystretch{1}
\begin{tabular}{c|c|c|cccc}
\hline
\multirow{2}{*}{\textbf{Method}} & \multirow{2}{*}{\textbf{Modal}} & \multirow{2}{*}{\scriptsize{\textbf{\#Par(M)}}} & \multicolumn{4}{c}{\textbf{Aquarium}} \\ \cline{4-7} 
 &  &  & \textbf{mAP} & \textbf{AP50} & \textbf{AP75} & \textbf{AR} \\ \hline
\multirow{3}{*}{\textbf{\begin{tabular}[c]{@{}c@{}}Learnable\\      linear projection\end{tabular}}} & \textbf{Image} & 13.82 & 51.7 & 81.0 & 55.2 & 65.5 \\
 & \textbf{Text} & 3.58 & 52.8 & 81.8 & 55.6 & 65.8 \\
 & \textbf{Dual} & 17.40 & 54.3 & 83.1 & 57.4 & 67.2 \\ \hline
\multirow{3}{*}{\textbf{\begin{tabular}[c]{@{}c@{}}Inverse\\      linear projection\end{tabular}}} & \textbf{Image} & \textbf{3.28} & 53.1 & 82.9 & 56.8 & 66.1 \\
 & \textbf{Text} & \textbf{3.28} & 55.4 & 83.6 & 58.7 & 68.1 \\
 & \textbf{Dual} & \textbf{3.28} & \textbf{57.7} & \textbf{85.0} & \textbf{60.6} & \textbf{70.5} \\ \hline
\end{tabular}
}
\end{minipage}
\end{table}

We conducted comprehensive ablation studies on SDPT. Unless otherwise specified, the dataset used is the Aquarium, and the employed VLPM is GLIP-L. Moreover, when performing ablations on a particular component, all other settings were kept consistent with optimal configurations. More ablation studies are provided in the supplementary materials.

\textbf{Synchronous dual-modal knowledge incorporation.} We conducted experiments to investigate the effectiveness of synchronous dual-modal knowledge incorporation, as shown in \cref{tab:synchrounous}. The term "Unshared token" describes a setup where distinct token sets are allocated to each modality, creating an asynchronous spatial configuration while keeping other settings unchanged. "Asynchronous training" also utilizes a shared set of unified prototype tokens but divides the training into two phases. Initially, unified prototype tokens are linked only to the text modality, and subsequently, in the second phase, to the image modality, introducing asynchronous learning over time. The findings confirm that SDPT, by synchronously integrating dual-modal knowledge both spatially and temporally, significantly improves tuning performance. More comparison results with learning separate modality-specific tokens for different modalities on LVIS are provided in the supplementary materials.

\textbf{Inverse linear projections.} We implemented a straightforward comparison by employing two learnable linear affine projections to transform $Z^i$ from the $d$ dimensional fusion space to the ${d}_{T}$ and ${d}_{I}$  dimensional latent spaces respectively, and compares with our inverse linear projections. Results are presented in \cref{tab:tuningfree}. In the table, "Modal" indicates the modal branch receiving the projected tokens; for example, "Text" implies the operation was solely applied to text tokens, and similarly for "Image". Results show that the inverse linear projections reduce the number of trainable parameters and markedly enhance performance. This underscores the benefit of maintaining the unbiased mapping distribution that was pre-trained on extensive data. In addition, the optimal result in the dual-modal further validates the effectiveness of dual-modal knowledge incorporation. 

\textbf{Unified prototype tokens' length.}
\label{sec:token length}
The length $k$ of unified prototype tokens is the only hyperparameter in SDPT that requires adjusting. We present the results of ablating $k$ in \cref{tab:tokenlength}. The table shows that even with $k=10$, SDPT achieves commendable performance, striking a good balance between lower fine-tuning costs and superior outcomes. Furthermore, as $k$ varies, SDPT maintains consistent effectiveness without drastic changes, resulting in smooth and regular performance curves. This confirms SDPT's robustness and the convenience of selecting the hyperparameter $k$. 


\begin{table}[t]
\centering 
\begin{minipage}[b]{0.43\linewidth} 
\centering 
\caption{Ablation on the length of unified prototype tokens, $k$. The best and second-best values are highlighted in \textbf{bold} and {\ul underlined}, respectively.}
\label{tab:tokenlength}
\resizebox{0.9\columnwidth}{!}{
\renewcommand\arraystretch{1}
\begin{tabular}{c|c|cccc}
\hline
\multirow{2}{*}{\textbf{$k$}} & \multirow{2}{*}{\scriptsize{\textbf{\#Par(M)}}} & \multicolumn{4}{c}{\textbf{Aquarium}} \\
 &  & \textbf{mAP} & \textbf{AP50} & \textbf{AP75} & \textbf{AR} \\ \hline
\textbf{10} & \textbf{0.16} & 57.1 & 84.5 & \textbf{61.9} & {\ul 69.5} \\
\textbf{100} & {\ul 1.64} & {\ul 57.6} & 84.7 & 63.4 & 69.4 \\
\textbf{200} & 3.28 & \textbf{57.7} & \textbf{85.0} & 60.6 & \textbf{70.5} \\
\textbf{400} & 6.56 & 57.5 & 84.7 & 61.1 & {\ul 69.5} \\
\textbf{800} & 13.11 & 57.1 & {\ul 84.8} & 58.6 & 68.7 \\
\textbf{1600} & 26.23 & 57.4 & 84.7 & {\ul 61.6} & {\ul 69.5} \\ \hline
\end{tabular}
}
\end{minipage}
\hfill 
\begin{minipage}[b]{0.51\linewidth} 
\centering 
\caption{Ablation on the position of the unified prototype tokens. "i→j" indicates the X-MHA layer indices where unified prototype tokens $Z^{i}$ are inserted. The best values are highlighted in \textbf{bold}.}
\label{tab:position}
\resizebox{0.65\columnwidth}{!}{
\renewcommand\arraystretch{1}
\begin{tabular}{c|c|c}
\hline
\multirow{2}{*}{\textbf{Layers}} & \multirow{2}{*}{\textbf{\scriptsize{\#Par(M)}}} & \textbf{Aquarium} \\
 &  & \textbf{mAP} \\ \hline
\textbf{1} & \textbf{0.41} & 57.3 \\
\textbf{1→4} & 1.64 & 57.1 \\
\textbf{1→8} & 3.28 & \textbf{57.7} \\
\textbf{5→8} & 1.64 & 57.4 \\
\textbf{8} & \textbf{0.41} & 57.5 \\ \hline
\end{tabular}
}
\end{minipage}
\end{table}

\textbf{Unified prototype tokens' position.}
\cref{tab:position} ablates which and how many layers to insert prompts. "i → j" indicates the X-MHA layer indices where learnable unified prototype tokens $Z^{i}$ are inserted. The notation "1" refers to the X-MHA layer closest to the input. GLIP-L has a total of 8 X-MHA layers. Overall, adding unified prototype tokens to all X-MHA layers yields the best performance. However, inserting tokens only in the first and eighth layer also attains commendable tuning results, demonstrating an excellent balance between performance and the costs associated with model training and storage.


\section{Conclusion}
In this paper, we propose Synchronous Dual Prompt Tuning (SDPT) tailored for fusion-based VLPMs. SDPT refines input features from dual modalities, aligning seamlessly with the intrinsic properties and transfer learning needs of fusion-based VLPMs. It utilizes unified prototype tokens $Z$ within the established fusion space and the inverse linear projections to maintain the integrity of pre-trained text-image aligning knowledge. This strategy circumvents learning inaccurate mapping or aligning distribution for downstream tasks. Furthermore, the two components enhance tuning efficiency with fewer parameters on fusion-based VLPMs. We conducted extensive experiments to validate the effectiveness of SDPT under various scenarios, demonstrating its generality and flexibility. Theoretically, SDPT could extend to other downstream tasks related to fusion-based VLPMs, such as phrase grounding, region-conditioned image generation \cite{li2023gligen}, etc. The limitation of this study lies in that it has not yet explored the effectiveness of SDPT on these tasks. These tasks will be a focal point for future work.

\subsubsection*{Acknowledgements} This work is supported by the National Natural Science Foundation in China under Grant 62371016 and U23B2063, the Bejing Natural Science Foundation Haidian District Joint Fund in China under Grant L222032.

\clearpage
\renewcommand\thesection{\Alph{section}} 
\setcounter{section}{0}
\section{Detailed Implementation}
\label{sec:imp}


We conducted the main experiments using the pre-trained GLIP-L model since GLIP \cite{li2022grounded} is the most representative fusion-based VLPMs. We adhered to the evaluation metrics outlined in the original GLIP paper, incorporating inference FLOPs and trainable parameters as criteria for efficient training. Unless specified otherwise, we used the following experimental settings. In GLIP-L, each fusion layer was augmented with the unified prototype tokens, each of a length denoted by $k$. For each benchmark, we chose different values of $k$ after conducting 10-fold cross-validation. To be noted that, due to ODinW13's task complexity, an aggressive cross-validation strategy was employed on one of its sub-datasets, Aquarium, to determine the hyperparameter $k$. The $1^{st}$ and $2^{nd}$ columns in Table.1 of the main text show values determined by the $k$ value confirmed on COCO's validation set \cite{lin2014microsoft}, serving as an example. We provide the determined $k$ values for COCO \cite{lin2014microsoft}, LVIS \cite{gupta2019lvis}, and ODinW13 \cite{li2022elevater} in \cref{stab:k} of the supplementary material. We used uniform random initialization between -1 and 1 for the unified prototype tokens. Regarding other settings, we followed the full fine-tuning settings presented in the original GLIP paper, including learning rate, training epochs, optimizer, etc.

\begin{table}[h]
\caption{$k$ values for COCO, LVIS, and ODinW13.}
 \vskip -0.23in
\label{stab:k}
\begin{center}
\resizebox{0.45\columnwidth}{!}{
\renewcommand\arraystretch{1}
\begin{tabular}{ccc}
\hline
\textbf{COCO\cite{lin2014microsoft}} & \textbf{LVIS\cite{gupta2019lvis}} & \textbf{ODinW13\cite{li2022elevater}} \\ \hline
$k$=120 & $k$=120 & $k$=200 \\ \hline
\end{tabular}
}
\end{center}
 \vskip -0.23in
\end{table}

Regarding comparison methods, since they are not specifically designed for fusion-based VLPMs, we had their hyperparameters optimally adjusted through cross-validation on COCO for GLIP-L, and applied these hyperparameters on other settings. Specific hyperparameter details for the comparison PEFT methods are provided in \cref{stab:para}. Implementation details for some comparison methods need to be specified. "LoRA" \cite{hu2021lora} and "BitFit" \cite{zaken2021bitfit} were applied exclusively to the dual encoders' stems for PEFT operations, whereas "LoRA$_{XMHA}$" and "BitFit$_{XMHA}$" were specifically targeted at X-MHA layers. "LoRA$_{XMHA}$" replaces all linear projections inside the X-MHA by LoRA linear projections. UPT designs self-attention structures to map dual-modality tokens which are then split and assigned to text and image latent spaces \cite{zang2022unified}. However, the token-wise split proposed in its original publication is impractical because the text and image latent spaces have different dimensions. Thus, UPT was reimplemented by channel-wise split due to the unavailability of the source code and the impractical token-wise split in its original paper. Other configurations not explicitly mentioned, such as insertion positions, were kept at their default settings. 

\begin{table}[t]
\centering 
\begin{minipage}[b]{0.495\linewidth} 
\centering 
\caption{Hyperparameters for the comparison PEFT methods.}
\label{stab:para}
\resizebox{\columnwidth}{!}{
\renewcommand\arraystretch{1}
\begin{tabular}{cc|c}
\hline
\multicolumn{2}{c|}{\textbf{Method}}                                            & \textbf{Hyperparameter}                         \\ \hline
\multicolumn{1}{c|}{\multirow{2}{*}{\textbf{Image}}} & \textbf{Adaptformer\cite{chen2022adaptformer}}     & middle dimension $\hat{d} = 800$ \\
\multicolumn{1}{c|}{}                                & \textbf{VPT\cite{jia2022visual}}             & prompt length $k=256$                           \\ \hline
\multicolumn{1}{c|}{\multirow{2}{*}{\textbf{Text}}}  & \textbf{Adapter\cite{houlsby2019parameter}}         & middle dimension $m=768$                        \\
\multicolumn{1}{c|}{}                                & \textbf{CoOp\cite{zhou2022learning}}            & context length $M=4$                           \\ \hline
\multicolumn{1}{c|}{\multirow{9}{*}{\textbf{Dual}}}  & \textbf{LoRA\cite{hu2021lora}}            & LoRA rank $r=16$                                  \\
\multicolumn{1}{c|}{}                                & \textbf{BitFit\cite{zaken2021bitfit}}          & fine-tune all bias parameters on encoders' stems \\
\multicolumn{1}{c|}{}                                & \textbf{LoRA$_{XMHA}$\cite{hu2021lora}}   & LoRA rank $r=16$                                  \\
\multicolumn{1}{c|}{}                                & \textbf{BitFit$_{XMHA}$\cite{zaken2021bitfit}} & fine-tune all bias parameters on X-MHA layers \\
\multicolumn{1}{c|}{}                                & \textbf{DPT\cite{xing2023dual}}             & the length of class-aware visual prompts = 20 \\
\multicolumn{1}{c|}{}                                & \textbf{Apollo\cite{chowdhury2023apollo}}          & prompt length = 192                             \\
\multicolumn{1}{c|}{}                                & \textbf{PMF\cite{li2023efficient}}             & prompt length $M=4$, starting fusion layer $L_{f}=4$  \\
\multicolumn{1}{c|}{}                                & \textbf{UPT\cite{zang2022unified}}             & prompt length $n=192$   \\
\multicolumn{1}{c|}{}                                & \textbf{MaPLe\cite{khattak2023maple}}           & prompt length = 192                    \\ \hline
\end{tabular}
}
\end{minipage}
\hfill 
\begin{minipage}[b]{0.48\linewidth} 
\centering 
\caption{The objects of interest for each subset and the image number of each split in ODinW13.}
\label{stab:odinwsetting}
\resizebox{\columnwidth}{!}{
\renewcommand\arraystretch{1}
\begin{tabular}{c|cccc}
\hline
\textbf{Dataset} & \textbf{Objects of interest} & \textbf{Train} & \textbf{Val} & \textbf{Test} \\ \hline
PascalVOC & Common objects (PascalVOC 2012) & 13690 & 3422 & \textbackslash{} \\
AerialDrone & Boats, cars, etc. from drone images & 52 & 15 & 7 \\
Aquarium & Penguins, starfish, etc. in an aquarium & 448 & 127 & 63 \\
Rabbits & Cottontail rabbits & 1980 & 19 & 10 \\
EgoHands & Hands in ego-centric images & 3840 & 480 & 480 \\
Mushrooms & Two kinds of mushrooms & 41 & 5 & 5 \\
Packages & Delivery packages & 19 & 4 & 3 \\
Raccoon & Raccoon & 150 & 29 & 17 \\
Shellfish & Shrimp, lobster, and crab & 406 & 116 & 58 \\
Vehicles & Car, bus, motorcycle, truck, and ambulance & 878 & 250 & 126 \\
Pistols & Pistol & 2377 & 297 & 297 \\
Pothole & Potholes on the road & 465 & 133 & 67 \\
Thermal & Dogs and people in thermal images & 142 & 41 & 20 \\ \hline
\end{tabular}
}
\end{minipage}
\end{table}

\subsection{Dataset details and splits}
COCO serves as a well-established benchmark for object detection. LVIS is a challenging dataset characterized by a long-tail distribution. We report results on LVIS MiniVal following MDETR \cite{kamath2021mdetr}. The OdinW13 benchmark presents a more challenging task for evaluating model performance in real-world scenarios, comprising 13 distinct subsets and demonstrating SDPT's practical transferability to downstream tasks. We adhered to the data splits in the original GLIP paper \cite{li2022grounded}. We reported results on COCO 2017val following \cite{lin2014microsoft}. We reported results on LVIS MiniVal following MDETR \cite{kamath2021mdetr}. In evaluating the transfer performance of our SDPT for fusion-based VLPMs on ODinW13, we adhered to the protocols established in the GLIP paper \cite{li2022grounded}. The specifics of OdinW13 are detailed in \cref{stab:odinwsetting}. For PascalVOC, we followed GLIP's approach and reported on the validation set.

\subsection{Details of self-training settings}

As for the self-training mode, we followed the standard self-training methodology described in \cite{dopido2013semisupervised}. We first utilized GLIP-L to generate the initial results which served as pseudo labels for self-training. The maximum of pseudo seeds was set as 20 per image. The other configurations were kept consistent with the main experiments.

\section{Comparison with straightforward combinations of single-modal methods}

We also trivially combined single-modality methods of different modalities to explore the effects of such straightforward combinations in dual-modality knowledge incorporation. The hyperparameters were kept consistent with those shown in \cref{stab:para} . The results are shown in \cref{stab:combination}. The superior results on ODinW13, listed in the final row of the table, were achieved with $k=200$, whereas COCO and LVIS employed $k=120$. The inference FLOPs and the number of trainable parameters for the final row were reported under $k=120$. 

\begin{table}[h]
\caption{Comparison with straightforward combinations of single-modal methods. "\#Par" denotes the number of trainable parameters. The values in the ODinW13 column represent the average mAP across 13 subtasks. Our method, SDPT, is highlighted with a gray background. The best and second-best values are highlighted in \textbf{bold} and {\ul underlined}, respectively.}
 \vskip -0.23in
\label{stab:combination}
\begin{center}
\resizebox{\columnwidth}{!}{
\renewcommand\arraystretch{1}
\begin{tabular}{c|cc|cccc|cccc|c}
\hline
 &  &  & \multicolumn{4}{c|}{\textbf{COCO}} & \multicolumn{4}{c|}{\textbf{LVIS}} & \textbf{ODinW13} \\
\multirow{-2}{*}{\textbf{Methods}} & \multirow{-2}{*}{\scriptsize{\textbf{FLOPs(G)}}} & \multirow{-2}{*}{\scriptsize{\textbf{\#Par(M)}}} & \textbf{mAP} & \textbf{AP50} & \textbf{AP75} & \textbf{AR} & \textbf{AP} & \textbf{APr} & \textbf{APc} & \textbf{APf} & \textbf{Full-shot} \\ \hline
\textbf{Adaptformer+Adapter} & 780.88 & 67.22 & 56.1 & 73.6 & 61.3 & 71.7 & 40.9 & 30.4 & 33.8 & 44.1 & 67.8 \\
\textbf{Adaptformer+CoOp} & 1362.81 & 57.13 & 55.9 & 73.2 & 61.2 & 71.4 & 40.4 & 30.1 & 34.4 & 43.6 & 66.9 \\
\textbf{VPT+Adapter} & 1317.42 & 14.88 & 56.6 & 74.1 & 61.8 & 72.5 & 41.1 & 30.7 & 34.5 & 44.4 & 68.1 \\
\textbf{VPT+CoOp} & 1910.51 & 4.78 & 56.2 & 73.3 & 61.7 & 72.3 & 40.4 & 29.3 & 32.5 & 41.4 & 68.7 \\
\rowcolor[HTML]{D9D9D9} 
\textbf{SDPT:k=10} & \textbf{724.77} & \textbf{0.16} & {\ul 57.6} & {\ul 75.4} & {\ul 63.0} & {\ul 72.8} & {\ul 41.2} & {\ul 31.5} & {\ul 34.9} & {\ul 45.0} & {\ul 69.5} \\
\rowcolor[HTML]{D9D9D9} 
\textbf{SDPT:k=120} & {\ul 738.36} & {\ul 1.97} & \textbf{58.0} & \textbf{75.8} & \textbf{63.2} & \textbf{73.1} & \textbf{41.4} & \textbf{31.8} & \textbf{35.2} & \textbf{45.1} & \textbf{71.2} \\ \hline
\end{tabular}
}
\end{center}
 \vskip -0.23in
\end{table}

\begin{figure}[h]
\centering
\includegraphics[width=0.5\columnwidth]{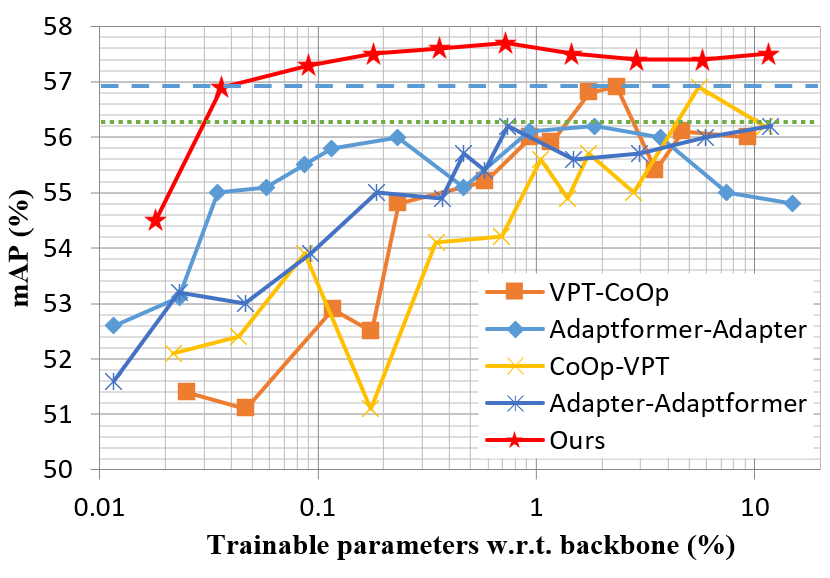}
  \caption{The optimization curve for hyperparameters, compared with straightforward combinations of single-modal methods. The blue dash line is the optimal score of "CoOp-VPT" and "VPT-CoOp" while the green dot line is the optimal score of "Adapter-Adaptformer" and "Adaptformer-Adapter".} 
  \label{opt_curve}
\vskip -0.205in
\end{figure}

In fact, when applying single-modal PEFT methods on fusion-based VLPMs, at least two individual hyperparameters (the hidden layer size for network-related methods and the token length for token-related methods) must be chosen. Illustrated in \cref{opt_curve}, the x-axis represents trainable parameters W.R.T backbone, and the y-axis represents mAP. "CoOp-VPT" indicates the change in performance when adjusting VPT hyperparameters after finding the optimal CoOp hyperparameters for the text modality, while "VPT-CoOp" denotes the same for the image modality. Similarly, the other two curves represent network-related methods. These curves exhibit significant fluctuation, suggesting the challenge of choosing optimal hyperparameter combinations. Our SDPT not only has one hyperparameter $k$ to be set but also produces smoother and more regular curves, making it easier to train good models.

\section{More Ablation Studies}

Unless otherwise specified, when performing ablations on a particular component, all other settings were kept consistent with optimal configurations.

\subsection{Comparison with learning separate modality-specific tokens for different modalities on LVIS} Introducing separate tokens is effective for capturing modality-specific knowledge in conventional VLPMs like CLIP but is unsuitable for fusion-based VLPMs, where multi-layer cross multi-head attentions (X-MHA) are deeply involved in pre-training. Modality-specific tokens require extra modal alignment with limited downstream data, potentially damaging the pre-trained generalized X-MHA knowledge. As shown in \cref{separate}, we compared the results on LVIS \cite{gupta2019lvis} of using separate modality-specific tokens for different branches (with token length matching our unified tokens and dimension matching the input tokens of each branch) and our method combined with separate tokens, on the fusion-based GLIP-L. Setting different tokens for each modality (Rows 1 and 3) reduces transfer generalization. In contrast, our unified prototype tokens, aided by inverse linear projections, better preserve the pre-trained knowledge and generalization of fusion-based VLPMs, achieving superior tuning performance with minimal trainable parameters.

\begin{table}[t]
\caption{Results of using separate modality-specific tokens on LVIS.}
\centering
\label{separate}
\resizebox{0.5\columnwidth}{!}{
\begin{tabular}{cccccc}
\hline
\textbf{Token type} & \textbf{\#Par(M)} & \textbf{AP} & \textbf{APr} & \textbf{APc} & \textbf{APf} \\ \hline
\textbf{Separate} & 16.81 & 38.9 & 29.9 & 34 & 43.1 \\
\textbf{Unified (Ours)} & \textbf{1.97} & \textbf{41.4} & \textbf{31.8} & \textbf{35.2} & \textbf{45.1} \\
\textbf{Unified + Separate} & 18.78 & 41.2 & 31.6 & 35.1 & 44.9 \\ \hline
\end{tabular}
}
\end{table}
\begin{table}[t]
\centering 
\begin{minipage}[b]{0.45\linewidth} 
\centering 
\caption{Effect of unified prototype tokens $Z$'s self-similarity on LVIS. The best values are highlighted in \textbf{bold}.}
\label{tab:self}
\resizebox{0.78\columnwidth}{!}{
\renewcommand\arraystretch{1}
\begin{tabular}{ccccc}
\hline
\textbf{Self-similarity of $Z$} & \textbf{AP} & \textbf{APr} & \textbf{APc} & \textbf{APf} \\ \hline
\textbf{Mask out} & 40.7 & 30.8 & 34.3 & 44.2 \\
\textbf{Retain (Ours)} & \textbf{41.4} & \textbf{31.8} & \textbf{35.2} & \textbf{45.1} \\ \hline
\end{tabular}
}
\end{minipage}
\hfill 
\begin{minipage}[b]{0.5\linewidth} 
\centering 
\caption{Adopting different initialization methods on LVIS. The best values are highlighted in \textbf{bold}.}
\label{tab:init}
\resizebox{0.96\columnwidth}{!}{
\renewcommand\arraystretch{1}
\begin{tabular}{ccccc}
\hline
\textbf{\begin{tabular}[c]{@{}c@{}}Uniform(-1,1)\\      (adopted)\end{tabular}} & \textbf{\begin{tabular}[c]{@{}c@{}}Standard\\      Normal\end{tabular}} & \textbf{Constant=0} & \textbf{\begin{tabular}[c]{@{}c@{}}Xavier\\      Normal\end{tabular}} & \textbf{\begin{tabular}[c]{@{}c@{}}Kaiming\\      Normal\end{tabular}} \\ \hline
41.4 & 40.6 & 39.9 & \textbf{41.5} & 41.4 \\ \hline
\end{tabular}
}
\end{minipage}
\end{table}

\subsection{Self-similarity of unified prototype tokens in cross-attention}
\cref{tab:self} presents the results of masking out the self-similarity of unified prototype tokens $Z$ during cross-attention on LVIS dataset with GLIP-L. It shows that masking out self-similarity performs a little bit worse than retaining. This implies that during cross-attention, the attention between $Z$ and text/image tokens is crucial for SDPT to model shared knowledge across domains, while calculating the self-similarity of $Z$ also benefits the process. 

\subsection{Initialization of unified prototype tokens}
We used uniform random initialization between -1 and 1 for the unified prototype tokens in the main experiments. \cref{tab:init} presents results (AP) on LVIS dataset with different initialization methods using GLIP-L, showing that SDPT performs well across various initialization strategies, demonstrating its stability.

\clearpage

\section{Detailed full-shot and few-shot results on ODinW13}
Detailed full-shot and few-shot comparisons are presented in \cref{stab:full-shot,stab:10-shot,stab:5-shot,stab:3-shot,stab:1-shot}. In the X-shot scenarios, we randomly sampled the dataset such that there are X examples per category \cite{kang2019few}. We changed the random seeds (and thus changed the sampled data) and conducted 3 independent runs for each X-shot experiment. The outcomes across these tables demonstrate that, within few-shot settings, SDPT consistently exhibits more stable performance across various downstream tasks.

\begin{table}[h]
\caption{Detailed full-shot results on ODinW13. Our method, SDPT, is highlighted with a gray background. The best values are highlighted in \textbf{bold}.}
 \vskip -0.23in
\label{stab:full-shot}
\begin{center}
\resizebox{\columnwidth}{!}{
\renewcommand\arraystretch{1}
\begin{tabular}{c|cccccccccccccc}
\hline
\textbf{Method} & \scriptsize{\textbf{AerialDrone}} & \scriptsize{\textbf{Vehicles}} & \scriptsize{\textbf{Aquarium}} & \scriptsize{\textbf{Mushrooms}} & \scriptsize{\textbf{Raccoon}} & \scriptsize{\textbf{Packages}} & \scriptsize{\textbf{Pothole}} & \scriptsize{\textbf{Shellfish}} & \scriptsize{\textbf{Rabbits}} & \scriptsize{\textbf{Pistols}} & \scriptsize{\textbf{Egohands}} & \scriptsize{\textbf{Pascalvoc}} & \scriptsize{\textbf{Thermal}} & \scriptsize{\textbf{Avg}} \\ \hline
\textbf{Linear probing} & 9.6 & 69.7 & 42.3 & 39.4 & 71.6 & 69.3 & 33.2 & \textbf{73.9} & 75.3 & 72.1 & 70.5 & 70.9 & 72.3 & 59.2 \\
\textbf{Full FT} & 32.6 & 71.6 & 56.6 & 88.1 & 69.4 & 67.1 & \textbf{60.3} & 65.8 & 76.4 & 75.7 & 79.4 & 69.6 & 83.1 & 68.9 \\
\textbf{Adaptformer} & 50.0 & 62.1 & 56.4 & 90.4 & 69.8 & 72.0 & 55.4 & 57.0 & 76.1 & 75.4 & 77.2 & 58.6 & 83.7 & 68.0 \\
\textbf{VPT} & 45.9 & 64.3 & 54.4 & 90.5 & 70.0 & 73.5 & 56.3 & 57.6 & 77.4 & 76.9 & \textbf{80.5} & 65.9 & 81.9 & 68.8 \\
\textbf{Adapter} & 50.9 & 63.4 & 57.3 & \textbf{91.1} & 72.2 & 75.7 & 56.4 & 58.4 & 78.1 & 78.1 & 77.6 & 58.6 & 82.2 & 69.2 \\
\textbf{CoOp} & 51.1 & 64.1 & 56.6 & 88.0 & 67.8 & 75.8 & 55.5 & 57.2 & 78.3 & 74.3 & 78.3 & 62.5 & 84.5 & 68.8 \\
\textbf{LoRA} & 50.5 & 64.2 & 56.4 & 88.5 & 70.2 & 73.8 & 55.1 & 59.7 & 77.9 & 76.4 & 80.1 & 59.8 & 81.6 & 66.9 \\
\textbf{BitFit} & 42.3 & 63.5 & 56.8 & 90.3 & 70.6 & 72.2 & 56.5 & 59.9 & 77.1 & 77.5 & 80.2 & 56.4 & 84.2 & 68.3 \\
\textbf{LoRA$_{XMHA}$} & 49.1 & 64.1 & 57.0 & 90.5 & 69.5 & \textbf{78.2} & 55.6 & 59.2 & 78.7 & \textbf{78.9} & 77.5 & 58.2 & 74.8 & 68.6 \\
\textbf{BitFit$_{XMHA}$} & 13.9 & \textbf{71.7} & 46.8 & 80.4 & 69.8 & 71.3 & 49.7 & 65.2 & 71.4 & 78.7 & 77.5 & 71.1 & 80.5 & 65.2 \\
\textbf{DPT} & 49.1 & 59.5 & 52.8 & 86.8 & 64.7 & 71.6 & 52.5 & 55.2 & 73.3 & 71.3 & 73.9 & 59.9 & 76.5 & 65.2 \\
\textbf{Apollo} & 47.5 & 61.3 & 56.5 & 91.0 & 65.9 & 73.4 & 52.9 & 61.5 & 78.4 & 74.7 & 79.8 & 56.9 & 80.6 & 67.7 \\
\textbf{PMF} & 51.2 & 63.5 & 56.7 & 91.0 & 71.0 & 68.0 & 56.1 & 59.5 & 78.3 & 77.4 & 79.2 & 57.6 & 83.3 & 68.7 \\
\textbf{UPT} & 50.9 & 61.8 & 55.8 & 90.4 & 70.7 & 66.8 & 52.1 & 55.4 & 74.5 & 72.8 & 73.5 & 59.4 & 82.7 & 66.7 \\
\textbf{MaPLe} & 48.8 & 62.5 & 56.3 & 88.4 & 71.4 & 72.1 & 53.1 & 58.0 & 77.4 & 73.6 & 76.9 & 63.4 & 81.4 & 67.4 \\
\rowcolor[HTML]{D9D9D9} 
\textbf{SDPT:k=10} & 50.6 & 66.1 & 57.1 & 90.5 & 69.1 & 76.4 & 56.0 & 61.4 & \textbf{79.3} & 77.0 & 78.3 & 59.8 & 81.9 & 69.5 \\
\rowcolor[HTML]{D9D9D9} 
\textbf{SDPT:k=200} & \textbf{52.0} & 64.5 & \textbf{57.7} & \textbf{91.1} & \textbf{72.5} & 76.2 & 57.4 & 60.4 & 78.9 & 78.5 & 80.1 & \textbf{71.4} & \textbf{84.7} & \textbf{71.2} \\ \hline
\end{tabular}
}
\end{center}
\vskip -0.23in
\end{table}
\begin{table}[h]
\caption{Detailed 10-shot results on ODinW13. Our method, SDPT, is highlighted with a gray background. The best values are highlighted in \textbf{bold}.}
 \vskip -0.23in
\label{stab:10-shot}
\begin{center}
\resizebox{\columnwidth}{!}{
\renewcommand\arraystretch{1}
\begin{tabular}{c|cccccccccccccc}
\hline
\textbf{Method} & \scriptsize{\textbf{AerialDrone}} & \scriptsize{\textbf{Vehicles}} & \scriptsize{\textbf{Aquarium}} & \scriptsize{\textbf{Mushrooms}} & \scriptsize{\textbf{Raccoon}} & \scriptsize{\textbf{Packages}} & \scriptsize{\textbf{Pothole}} & \scriptsize{\textbf{Shellfish}} & \scriptsize{\textbf{Rabbits}} & \scriptsize{\textbf{Pistols}} & \scriptsize{\textbf{Egohands}} & \scriptsize{\textbf{Pascalvoc}} & \scriptsize{\textbf{Thermal}} & \scriptsize{\textbf{Avg}} \\ \hline
\textbf{Linear probing} & 11.5 & 65.6 & 35.1 & 38.0 & 66.7 & 71.7 & 25.8 & 72.5 & 74.0 & 67.9 & 64.7 & 65.2 & 67.2 & 55.8 \\
\textbf{Full FT} & 32.0 & 71.5 & 52.3 & 88.1 & 64.7 & 67.1 & \textbf{44.3} & \textbf{69.4} & 70.6 & 68.4 & 72.4 & 66.4 & 76.3 & 64.9 \\
\textbf{Adaptformer} & 40.4 & 54.0 & 55.7 & 89.4 & 62.8 & 72.7 & 33.4 & 40.9 & 72.7 & 57.4 & \textbf{78.9} & 54.5 & 68.4 & 60.1 \\
\textbf{VPT} & 42.5 & 66.1 & 53.8 & \textbf{92.8} & 69.3 & 75.9 & 38.7 & 51.9 & 76.2 & 72.5 & 72.3 & 59.6 & 75.8 & 65.2 \\
\textbf{Adapter} & 43.8 & 66.4 & 55.7 & 90.4 & \textbf{71.8} & 71.6 & 34.0 & 56.3 & \textbf{77.4} & 67.5 & 72.0 & 59.3 & 78.2 & 65.0 \\
\textbf{CoOp} & 44.0 & 66.5 & 54.6 & 91.6 & 70.7 & 73.5 & 41.5 & 57.8 & 73.0 & 70.4 & 72.0 & 55.9 & \textbf{79.9} & 65.5 \\
\textbf{LoRA} & 43.5 & 64.5 & 55.3 & 89.7 & 66.8 & 71.9 & 39.1 & 58.4 & 74.5 & 70.2 & 71.1 & 62.3 & 74.5 & 64.8 \\
\textbf{BitFit} & 42.3 & 68.1 & 49.0 & 90.1 & 67.9 & \textbf{77.8} & 38.8 & 57.5 & 74.6 & 72.7 & 71.8 & 65.0 & 77.5 & 65.6 \\
\textbf{LoRA$_{XMHA}$} & 44.4 & 62.9 & 43.1 & 89.4 & 63.4 & 71.3 & 40.8 & 58.7 & 72.6 & 71.0 & 71.3 & 61.2 & 73.2 & 63.3 \\
\textbf{BitFit$_{XMHA}$} & 30.7 & 68.3 & 45.4 & 85.2 & 67.1 & 72.3 & 32.9 & 64.3 & \textbf{77.4} & \textbf{75.3} & 72.8 & \textbf{66.7} & 67.4 & 63.5 \\
\textbf{DPT} & 42.1 & 63.4 & 52.8 & 83.4 & 66.1 & 70.1 & 33.5 & 51.1 & 69.7 & 68.4 & 70.2 & 58.8 & 71.5 & 61.6 \\
\textbf{Apollo} & \textbf{45.5} & 52.8 & 49.8 & 89.1 & 68.0 & 68.5 & 41.7 & 57.8 & 76.2 & 69.2 & 71.7 & 61.5 & 78.7 & 63.9 \\
\textbf{PMF} & 44.6 & 66.7 & 50.5 & 91.3 & 67.9 & 71.0 & 40.0 & 55.7 & 73.4 & 72.8 & 71.4 & 54.8 & 73.7 & 64.1 \\
\textbf{UPT} & 43.3 & 64.5 & 52.9 & 84.1 & 67.0 & 70.2 & 39.5 & 57.6 & 70.5 & 68.3 & 70.9 & 56.4 & 66.8 & 62.5 \\
\textbf{MaPLe} & 44.2 & 66.0 & 54.2 & 87.1 & 66.4 & 71.8 & 38.5 & 60.1 & 72.2 & 71.0 & 70.8 & 57.7 & 71.9 & 64.0 \\
\rowcolor[HTML]{D9D9D9} 
\textbf{SDPT:k=10} & 43.3 & \textbf{68.8} & 49.6 & \textbf{92.8} & 67.0 & 77.1 & 40.5 & 58.2 & 76.6 & 69.9 & 71.3 & 63.4 & 74.6 & 65.6 \\
\rowcolor[HTML]{D9D9D9} 
\textbf{SDPT:k=200} & 44.4 & 66.1 & \textbf{57.3} & 90.4 & 68.3 & 73.2 & 40.7 & 60.6 & 76.2 & 71.1 & 72.1 & 64.3 & 77.9 & \textbf{66.4} \\ \hline
\end{tabular}
}
\end{center}
\vskip -0.23in
\end{table}
\begin{table}[h]
\caption{Detailed 5-shot results on ODinW13. Our method, SDPT, is highlighted with a gray background. The best values are highlighted in \textbf{bold}.}
 \vskip -0.23in
\label{stab:5-shot}
\begin{center}
\resizebox{\columnwidth}{!}{
\renewcommand\arraystretch{1}
\begin{tabular}{c|cccccccccccccc}
\hline
\textbf{Method} & \scriptsize{\textbf{AerialDrone}} & \scriptsize{\textbf{Vehicles}} & \scriptsize{\textbf{Aquarium}} & \scriptsize{\textbf{Mushrooms}} & \scriptsize{\textbf{Raccoon}} & \scriptsize{\textbf{Packages}} & \scriptsize{\textbf{Pothole}} & \scriptsize{\textbf{Shellfish}} & \scriptsize{\textbf{Rabbits}} & \scriptsize{\textbf{Pistols}} & \scriptsize{\textbf{Egohands}} & \scriptsize{\textbf{Pascalvoc}} & \scriptsize{\textbf{Thermal}} & \scriptsize{\textbf{Avg}} \\ \hline
\textbf{Linear probing} & 8.8 & 64.2 & 33.4 & 37.2 & 69.2 & 69.3 & 25.3 & \textbf{71.5} & 74.1 & 68.0 & 63.8 & 65.0 & 65.2 & 55.0 \\
\textbf{Full FT} & 26.4 & \textbf{70.0} & 49.5 & 88.1 & 68.8 & 71.1 & 39.9 & 68.5 & 70.7 & 68.3 & 71.9 & 66.6 & \textbf{75.2} & 64.2 \\
\textbf{Adaptformer} & 37.7 & 48.2 & 54.7 & 87.8 & 64.1 & 73.4 & 35.3 & 33.3 & 71.8 & 46.7 & 64.7 & 50.6 & 60.7 & 56.1 \\
\textbf{VPT} & 39.1 & 59.7 & 52.6 & 90.4 & 65.9 & 69.0 & 33.4 & 59.4 & 74.3 & 70.8 & 70.5 & 60.1 & 70.6 & 63.6 \\
\textbf{Adapter} & 38.7 & 56.2 & 54.1 & 90.4 & \textbf{71.7} & 71.6 & 35.3 & 52.3 & 74.1 & 70.8 & 70.5 & 61.5 & 65.8 & 62.5 \\
\textbf{CoOp} & 39.4 & 64.1 & 52.8 & \textbf{91.6} & 68.9 & 71.5 & 36.0 & 60.9 & 72.2 & 70.4 & 71.3 & 56.5 & 65.2 & 63.1 \\
\textbf{LoRA} & 36.4 & 65.9 & 51.3 & 90.5 & 65.4 & 75.1 & 36.1 & 58.2 & 72.1 & 70.8 & 70.4 & 62.1 & 66.4 & 63.1 \\
\textbf{BitFit} & 36.2 & 66.5 & 45.0 & 92.2 & 67.4 & \textbf{75.2} & 33.0 & 62.4 & 76.6 & 72.2 & 70.8 & 63.1 & 73.1 & 64.1 \\
\textbf{LoRA$_{XMHA}$} & 37.6 & 65.9 & 47.4 & 87.8 & 68.4 & 69.5 & \textbf{42.3} & 54.6 & 71.8 & \textbf{73.2} & 70.3 & 62.1 & 70.3 & 63.2 \\
\textbf{BitFit$_{XMHA}$} & 27.4 & 68.1 & 43.5 & 85.1 & 68.5 & 72.5 & 34.1 & 64.1 & 76.4 & 64.9 & \textbf{72.0} & \textbf{67.2} & 73.9 & 62.9 \\
\textbf{DPT} & 31.4 & 67.3 & 44.9 & 84.8 & 67.0 & 70.2 & 35.1 & 53.4 & 72.3 & 69.6 & 67.4 & 56.7 & 68.1 & 60.6 \\
\textbf{Apollo} & \textbf{40.3} & 68.8 & 46.6 & 91.0 & 64.3 & 72.6 & 37.5 & 57.4 & 74.6 & 69.7 & 68.9 & 58.8 & 72.8 & 63.3 \\
\textbf{PMF} & 38.7 & 63.7 & 44.6 & 84.4 & 67.4 & 70.8 & 36.4 & 62.7 & 74.9 & 72.4 & 71.2 & 59.7 & 69.9 & 62.8 \\
\textbf{UPT} & 33.4 & 63.4 & 48.1 & 81.2 & 63.1 & 69.9 & 34.1 & 53.1 & 72.5 & 70.5 & 68.5 & 60.2 & 67.6 & 60.4 \\
\textbf{MaPLe} & 34.3 & 64.8 & 47.5 & 85.4 & 67.1 & 70.6 & 35.7 & 54.8 & 75.6 & 71.4 & 68.8 & 60.9 & 67.0 & 61.8 \\
\rowcolor[HTML]{D9D9D9} 
\textbf{SDPT:k=10} & 38.4 & 66.8 & 49.2 & 91.0 & 68.2 & 74.2 & 40.1 & 54.4 & \textbf{76.9} & 72.3 & 70.7 & 61.8 & 71.9 & 64.3 \\
\rowcolor[HTML]{D9D9D9} 
\textbf{SDPT:k=200} & 38.4 & 67.3 & \textbf{56.6} & \textbf{91.6} & 68.2 & 73.4 & 37.6 & 53.4 & 75.9 & 72.9 & 70.5 & 61.2 & 69.9 & \textbf{64.4} \\ \hline
\end{tabular}
}
\end{center}
\vskip -0.23in
\end{table}
\begin{table}[h]
\caption{Detailed 3-shot results on ODinW13. Our method, SDPT, is highlighted with a gray background. The best values are highlighted in \textbf{bold}.}
 \vskip -0.23in
\label{stab:3-shot}
\begin{center}
\resizebox{\columnwidth}{!}{
\renewcommand\arraystretch{1}
\begin{tabular}{c|cccccccccccccc}
\hline
\textbf{Method} & \scriptsize{\textbf{AerialDrone}} & \scriptsize{\textbf{Vehicles}} & \scriptsize{\textbf{Aquarium}} & \scriptsize{\textbf{Mushrooms}} & \scriptsize{\textbf{Raccoon}} & \scriptsize{\textbf{Packages}} & \scriptsize{\textbf{Pothole}} & \scriptsize{\textbf{Shellfish}} & \scriptsize{\textbf{Rabbits}} & \scriptsize{\textbf{Pistols}} & \scriptsize{\textbf{Egohands}} & \scriptsize{\textbf{Pascalvoc}} & \scriptsize{\textbf{Thermal}} & \scriptsize{\textbf{Avg}} \\ \hline
\textbf{Linear probing} & 8.5 & 63.2 & 33.7 & 37.0 & 66.6 & 69.3 & 24.8 & \textbf{71.2} & 74.3 & 68.0 & 64.1 & 64.8 & 65.9 & 54.7 \\
\textbf{Full FT} & 22.3 & 66.7 & 45.2 & 81.6 & 65.3 & 71.8 & 37.0 & 67.6 & 72.3 & 68.1 & 70.4 & 65.6 & 73.1 & 62.1 \\
\textbf{Adaptformer} & 35.7 & 48.4 & 46.4 & 90.3 & 62.8 & 69.8 & 20.1 & 41.9 & 70.4 & 53.3 & 63.1 & 45.6 & 67.6 & 55.0 \\
\textbf{VPT} & 38.2 & 63.2 & 45.1 & 85.9 & 62.5 & 71.5 & 31.5 & 54.2 & 77.6 & 66.6 & 70.9 & 59.6 & 69.5 & 61.3 \\
\textbf{Adapter} & 39.1 & 60.9 & 47.7 & 85.7 & 65.8 & 71.1 & 33.3 & 54.3 & 74.0 & 68.0 & 69.5 & 59.3 & 71.6 & 61.6 \\
\textbf{CoOp} & 39.2 & 61.8 & 45.8 & 89.6 & 64.8 & 71.7 & 33.6 & 54.4 & 76.2 & 64.1 & 70.2 & 61.0 & 68.9 & 61.6 \\
\textbf{LoRA} & 35.4 & 62.1 & 44.1 & 87.8 & 66.2 & 70.5 & 31.5 & 61.1 & 73.5 & 67.2 & 70.3 & 60.4 & 71.5 & 61.7 \\
\textbf{BitFit} & 32.3 & 62.5 & 43.7 & 87.3 & 64.9 & 67.7 & 31.9 & 62.3 & 71.2 & 68.9 & 71.1 & 64.4 & \textbf{73.4} & 61.7 \\
\textbf{LoRA$_{XMHA}$} & 38.0 & 56.9 & 43.1 & 89.4 & 63.4 & 71.3 & 32.5 & 58.7 & 72.6 & 71.0 & 71.3 & 55.6 & 73.2 & 61.3 \\
\textbf{BitFit$_{XMHA}$} & 27.4 & \textbf{66.8} & 39.3 & 84.4 & 68.1 & 70.4 & 31.7 & 58.5 & \textbf{77.7} & \textbf{74.1} & 71.3 & \textbf{65.9} & 72.1 & 62.1 \\
\textbf{DPT} & 31.0 & 66.2 & 41.3 & 84.1 & 66.4 & 68.7 & 35.0 & 51.9 & 72.2 & 68.4 & 66.3 & 56.7 & 67.3 & 59.7 \\
\textbf{Apollo} & 36.4 & 61.5 & 44.3 & 91.0 & 63.1 & \textbf{73.4} & 33.5 & 55.7 & 74.4 & 69.6 & 70.0 & 60.8 & 70.0 & 61.8 \\
\textbf{PMF} & 37.3 & 64.5 & 44.6 & 89.6 & 65.1 & \textbf{73.4} & 32.5 & 54.4 & 72.5 & 67.5 & \textbf{71.6} & 61.0 & 71.1 & 61.9 \\
\textbf{UPT} & 34.2 & 63.1 & 48.0 & 80.5 & 62.7 & 68.6 & 33.7 & 53.6 & 73.5 & 70.5 & 68.3 & 60.1 & 66.1 & 60.2 \\
\textbf{MaPLe} & 34.1 & 64.3 & 46.1 & 83.7 & 66.9 & 69.3 & 34.9 & 54.1 & 75.3 & 71.1 & 68.5 & 60.4 & 66.0 & 61.1 \\
\rowcolor[HTML]{D9D9D9} 
\textbf{SDPT:k=10} & 38.4 & 65.6 & 45.0 & 90.4 & 67.8 & 72.8 & 32.4 & 53.1 & 72.2 & 68.2 & 70.4 & 60.2 & 72.3 & 62.2 \\
\rowcolor[HTML]{D9D9D9} 
\textbf{SDPT:k=200} & \textbf{39.4} & 66.0 & \textbf{49.5} & \textbf{91.6} & \textbf{70.7} & 72.2 & \textbf{39.7} & 54.6 & 77.1 & 71.1 & 71.4 & 61.3 & 72.2 & \textbf{64.4} \\ \hline
\end{tabular}
}
\end{center}
 \vskip -0.23in
\end{table}
\begin{table}[h]
\caption{Detailed 1-shot results on ODinW13. Our method, SDPT, is highlighted with a gray background. The best values are highlighted in \textbf{bold}.}
 \vskip -0.23in
\label{stab:1-shot}
\begin{center}
\resizebox{\columnwidth}{!}{
\renewcommand\arraystretch{1}
\begin{tabular}{c|cccccccccccccc}
\hline
\textbf{Method} & \scriptsize{\textbf{AerialDrone}} & \scriptsize{\textbf{Vehicles}} & \scriptsize{\textbf{Aquarium}} & \scriptsize{\textbf{Mushrooms}} & \scriptsize{\textbf{Raccoon}} & \scriptsize{\textbf{Packages}} & \scriptsize{\textbf{Pothole}} & \scriptsize{\textbf{Shellfish}} & \scriptsize{\textbf{Rabbits}} & \scriptsize{\textbf{Pistols}} & \scriptsize{\textbf{Egohands}} & \scriptsize{\textbf{Pascalvoc}} & \scriptsize{\textbf{Thermal}} & \scriptsize{\textbf{Avg}} \\ \hline
\textbf{Linear probing} & 7.6 & 60.5 & 28.1 & 41.3 & 67.0 & 70.2 & 24.8 & \textbf{71.0} & 74.6 & 67.9 & 60.3 & 63.7 & 66.1 & 54.1 \\
\textbf{Full FT} & 18.7 & 65.4 & 39.5 & 69.8 & \textbf{68.4} & 70.6 & 28.9 & \textbf{71.0} & 70.0 & 68.1 & 70.5 & \textbf{64.8} & \textbf{72.9} & 59.9 \\
\textbf{Adaptformer} & 25.7 & 41.7 & 33.2 & 91.0 & 60.0 & 68.8 & 8.4 & 25.6 & 73.7 & 34.2 & 62.9 & 33.5 & 56.0 & 47.3 \\
\textbf{VPT} & 30.8 & 61.5 & 33.1 & 91.0 & 66.6 & 71.0 & 30.9 & 51.7 & 75.7 & 70.9 & 69.3 & 56.3 & 57.3 & 58.9 \\
\textbf{Adapter} & 34.3 & 65.5 & 32.9 & 89.9 & 66.3 & 69.3 & 27.9 & 60.7 & \textbf{78.4} & 67.1 & 70.6 & 56.3 & 58.3 & 59.8 \\
\textbf{CoOp} & 33.6 & 55.5 & 34.1 & 91.0 & 64.6 & 70.2 & 27.2 & 56.0 & 78.3 & 73.4 & 69.9 & 58.0 & 62.5 & 59.6 \\
\textbf{LoRA} & \textbf{34.9} & 62.0 & 41.5 & 86.1 & 64.8 & 70.2 & 30.3 & 60.4 & 72.1 & 67.6 & 70.1 & 60.1 & 71.2 & 60.9 \\
\textbf{BitFit} & 30.9 & 64.9 & 37.5 & 91.0 & 66.4 & \textbf{73.3} & 29.2 & 56.8 & 74.3 & 73.0 & 70.5 & 63.9 & 63.8 & 61.2 \\
\textbf{LoRA$_{XMHA}$} & 31.9 & 61.5 & 38.9 & 91.0 & 58.7 & 70.1 & 30.7 & 57.8 & 72.7 & 68.5 & \textbf{71.7} & 59.4 & 67.0 & 60.0 \\
\textbf{BitFit$_{XMHA}$} & 25.7 & 65.1 & 33.8 & 90.4 & 66.4 & 67.6 & 30.9 & 62.2 & 75.9 & \textbf{75.2} & 70.7 & \textbf{64.8} & 71.4 & 61.5 \\
\textbf{DPT} & 30.5 & 63.3 & \textbf{41.7} & 85.4 & 61.7 & 66.5 & 30.1 & 53.3 & 71.5 & 69.4 & 69.8 & 59.7 & 64.4 & 59.0 \\
\textbf{Apollo} & 32.7 & 62.8 & 40.8 & 79.6 & 65.2 & 68.7 & 30.2 & 57.6 & 73.6 & 70.6 & \textbf{71.7} & 59.0 & 61.7 & 59.6 \\
\textbf{PMF} & 32.9 & 66.7 & 39.3 & 88.4 & 67.8 & 69.9 & 28.2 & 60.2 & 75.0 & 70.0 & 71.4 & 58.0 & 62.8 & 60.8 \\
\textbf{UPT} & 30.1 & 61.0 & 33.6 & 80.4 & 60.4 & 63.6 & 28.5 & 50.5 & 69.4 & 65.8 & 66.1 & 55.6 & 60.9 & 55.8 \\
\textbf{MaPLe} & 32.1 & 63.4 & 38.3 & 82.1 & 62.7 & 64.4 & 30.1 & 53.7 & 72.5 & 71.0 & 67.8 & 60.1 & 63.5 & 58.6 \\
\rowcolor[HTML]{D9D9D9} 
\textbf{SDPT:k=10} & 33.0 & \textbf{68.5} & 38.1 & 88.7 & 65.5 & 70.7 & 29.5 & 56.7 & 76.0 & 72.4 & 71.1 & 60.5 & 68.6 & 61.5 \\
\rowcolor[HTML]{D9D9D9} 
\textbf{SDPT:k=200} & 33.8 & 66.2 & 37.2 & \textbf{91.6} & 64.7 & 71.3 & \textbf{31.4} & 58.7 & 77.8 & 72.0 & \textbf{71.7} & 61.0 & 63.0 & \textbf{61.6} \\ \hline
\end{tabular}
}
\end{center}
 \vskip -0.23in
\end{table}

\clearpage

\section{Additional Visualization Examples}

\begin{figure}[h]
    \centering
    \includegraphics[width=\columnwidth]{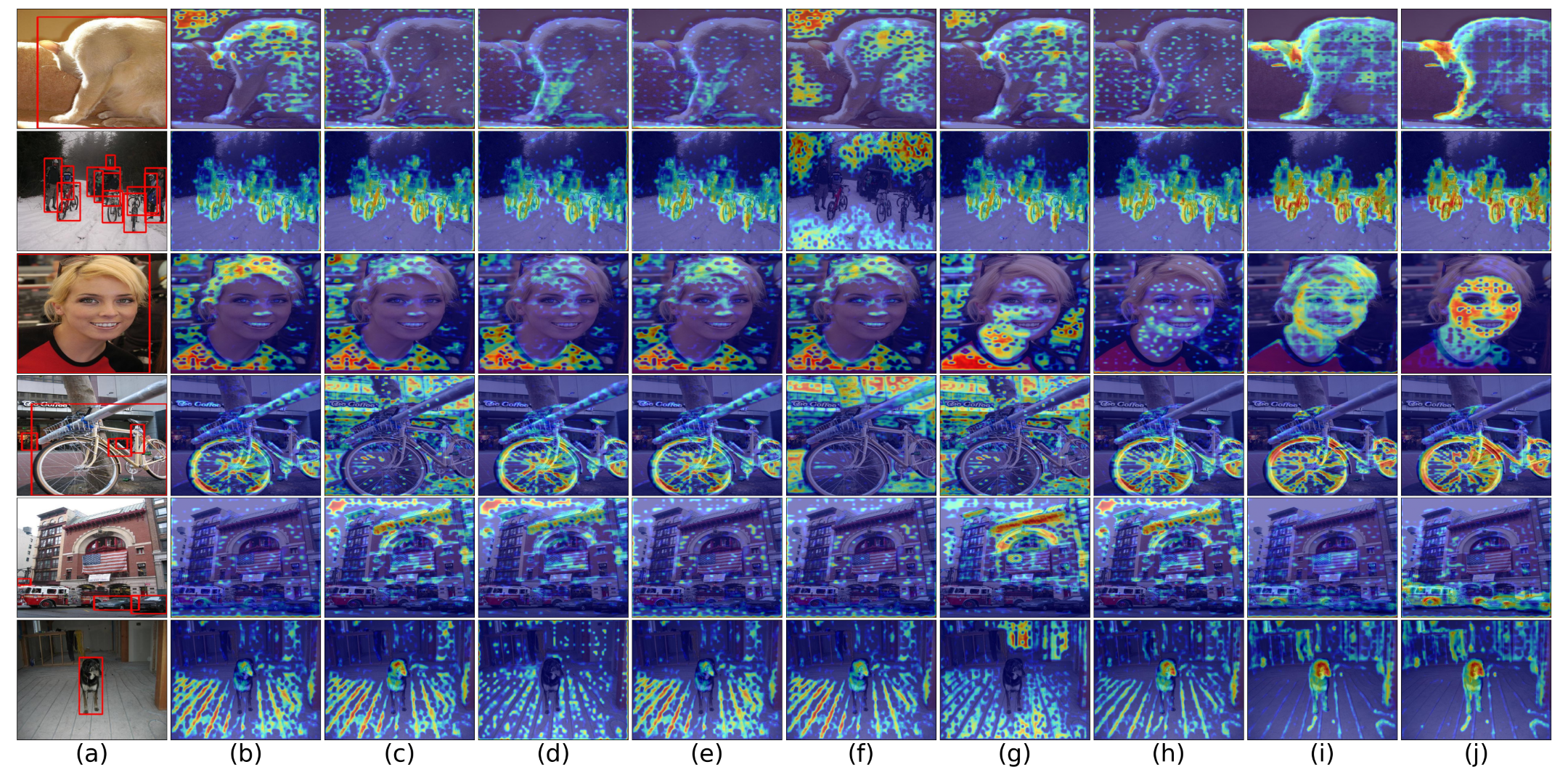}
    \caption{Comparison of attention map visualization of the different methods on COCO. (a) Original image and ground truths, (b) LoRA, (c) BitFit, (d) DPT, (e) Apollo, (f) PMF, (g) UPT, (h) MaPLe, (i) SDPT (k=10), (j) SDPT (k=120). Ground truths are marked by red boxes.} 
    \label{fig:coco}
\end{figure}

\begin{figure}[h]
    \centering
    \includegraphics[width=\columnwidth]{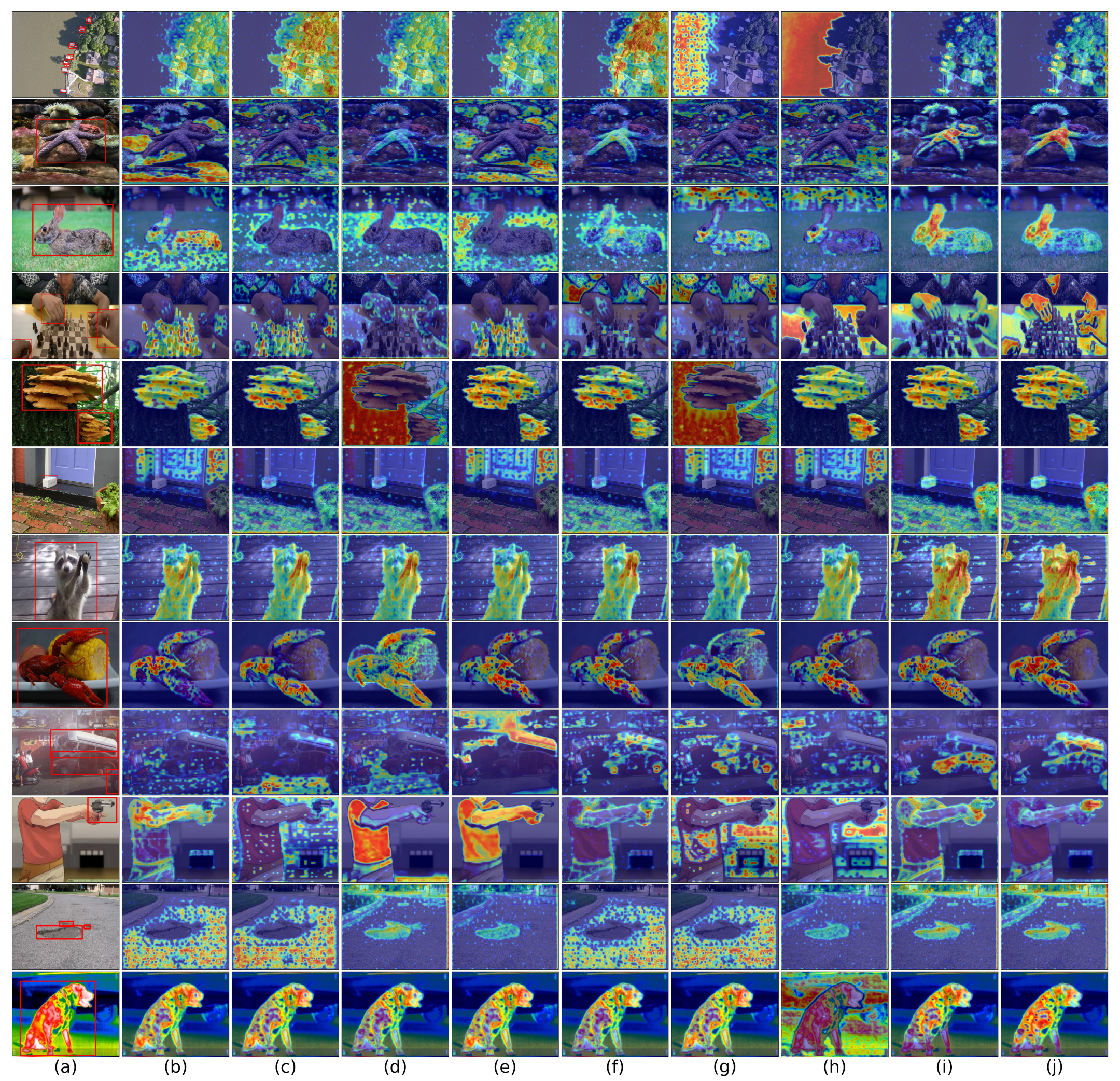}
    \caption{Comparison of attention map visualization of the different methods on ODinW13. From the top row to the bottom row: AerialDrone, Aquarium, Rabbits, EgoHands, Mushrooms, Packages, Raccoon, Shellfish, Vehicles, Pistols, Pothole, Thermal. (a) Original image and ground truths, (b) LoRA, (c) BitFit, (d) DPT, (e) Apollo, (f) PMF, (g) UPT, (h) MaPLe, (i) SDPT (k=10), (j) SDPT (k=200). Ground truths are marked by red boxes.} 
    \label{fig:odinw}
\end{figure}

In \cref{fig:coco,fig:odinw}, we present a series of attention heatmaps for category text, generated using the GLIP-L model fine-tuned on COCO, or ODinW13 excluding PascalVOC. In the case of comparison methods, we observed instances of attention dispersion, likely due to overfitting, resulting in a reduced focus on the target object. In stark contrast, our SDPT effectively directs the fine-tuned GLIP's attention more precisely onto the target object. This highlights the significant advantage of our approach in maximizing the efficacy of fusion-based VLPMs during the fine-tuning process.

\clearpage
%
%
\bibliographystyle{splncs04}
\bibliography{main}
\end{document}